\documentclass[
twocolumn,
superscriptaddress,
amsmath,
amssymb,
aps,
prb,
floatfix,
]{revtex4-2}

\usepackage[utf8]{inputenc}
\usepackage{graphicx}
\usepackage{dcolumn}
\usepackage{bm}
\usepackage{mathrsfs}
\usepackage{physics}
\usepackage{xcolor}
\usepackage{amsmath, amssymb, mathtools}
\usepackage[colorlinks=true]{hyperref}
\usepackage{listings}
\definecolor{jpblue}{RGB}{0,0,180}
\definecolor{jpgreen}{RGB}{0,128,0}
\definecolor{jpred}{RGB}{163,21,21}

\usepackage[T1]{fontenc}
\usepackage[utf8]{inputenc} 
\usepackage{textcomp}
\usepackage{lmodern}

\usepackage[frozencache,cachedir=.]{minted}

\usepackage{comment} 
\usepackage{orcidlink}
\usepackage{subcaption}

\definecolor{ve}{RGB}{0,161,22}
\definecolor{dmag}{rgb}{0.6,0.0,0.6}
\definecolor{pink}{rgb}{1,0,0.9}

\graphicspath{{figs/}} 

\begin{document}

\title{From Classical to Quantum Reinforcement Learning and Its Applications in Quantum Control: A Beginner's Tutorial}

\begin{abstract}
This tutorial is designed to make reinforcement learning (RL) more accessible to undergraduate students and researchers by offering clear, example-driven explanations. It focuses on bridging the gap between RL theory and practical coding, addressing common challenges that one faces when transitioning from conceptual understanding to implementation. Through hands-on examples and approachable explanations, the tutorial aims to equip the reader with the foundational skills needed to confidently apply RL techniques in real-world scenarios.
\textit{Code Availability}: \url{https://github.com/asen009/Reinforcement_Tutorial_Undergraduates}

\end{abstract}

\author{Abhijit Sen\orcidlink{0000-0003-2783-1763}}
\email{asen1@tulane.edu, abhijit913@gmail.com}
\affiliation{Department of Physics and Engineering Physics, Tulane University, New Orleans, LA 70118, USA}
\author{Sonali Panda \orcidlink{0009-0008-6027-8342}}
\email{sp.sonalpanda@gmail.com}
\affiliation{Department of Physics, Indian Institute of Technology,  Dhanbad, India}

\author{Mahima Arya \orcidlink{0000-0002-1847-9705}}
\email{marya@tulane.edu, aryamahima@gmail.com}
\affiliation{Department of Physics and Engineering Physics, Tulane University, New Orleans, LA 70118, USA}

\author{Subhajit Patra\orcidlink{0009-0001-6043-8747}}
\email{subh.p9083@gmail.com}
\affiliation{Department of Electrical Engineering, Indian Institute of Technology, Delhi, India}

\author{Zizhan Zheng\orcidlink{0000-0000-0000-0000}}  
\email{zzheng3@tulane.edu}
\affiliation{Department of Computer Science, Tulane University, New Orleans, LA 70118, USA}

\author{Denys I. Bondar\orcidlink{0000-0002-3626-4804}}
\email{dbondar@tulane.edu}
\affiliation{Department of Physics and Engineering Physics, Tulane University, New Orleans, LA 70118, USA}

\date{\today}

\maketitle

\twocolumngrid  

\section{Introduction}

Reinforcement Learning (RL) is a branch of artificial intelligence that trains an agent to maximize desirable outcomes through interaction with its environment. RL has seen rapid advancements in recent years, leading to breakthroughs in various fields. In the field of games \cite{Schrittwieser2020, Silver2018}, autonomous navigation \cite{Bellemare2020, Irshayyid2024}, robotics \cite{Khan2020} and Large Language Models~\cite{LLM}, significant development was possible due to RL. In the near future, AI models are expected to increasingly integrate supervised and unsupervised learning algorithms with RL, creating hybrid approaches that leverage the strengths of each method. It is advised to refer to the following resource as a prerequisite \cite{sen2025reinforcementtutorial}.

The study of RL is crucial for researchers due to its growing applications in various fields \cite{jaeger}. However, one notable shortcoming of the field is the significant time investment required to master its concepts and techniques. It is crucial to bridge the gap between theoretical understanding and practical code implementation in reinforcement learning, ensuring that learners grasp both the concepts and their real world application. 
Many tutorials and journal articles on reinforcement learning provide useful insights\cite{oldtutorial1, nutshell, oldtutorial3}, but they are often lengthy, mathematically heavy, or scattered across multiple examples, making it hard for readers to build a clear understanding. This tutorial takes a different approach by using a single, simple example to explain all the main concepts in a connected and easy-to-follow way. It focuses only on the ideas that are necessary and sufficient, keeping the content concise but complete. Each method is introduced by addressing what the previous one lacks, helping readers see how the techniques improve step by step. The tutorial also includes clear mathematical explanations along with ready-to-use code, making it practical and accessible for anyone looking to learn reinforcement learning in a structured and efficient manner.

Before explaining how RL can be used for quantum control, we will first present RL in simple yet complete terms. Hence, the rest of the paper is organized as follows: First of all in Sec.~\ref{Sec2}, we provide an overview of the fundamental concepts of RL. Then in ~\ref{Sec2.1}, we have explicitly discussed the concept of probability which is an essential prerequisite. The subsequent sections~\ref{Sec3} and \ref{Sec4} delve into the essential terminology commonly used in RL. In Sec.~\ref{Sec5}, we discuss key elements such as policies, reward systems, and actions. Further, we explore policy improvement techniques, while in Sec.~\ref{objective2} we continue the discussion on policy improvement strategies. Section~\ref{Sec7} covers the Markov Decision Process (MDP) and its associated techniques.  Dynamic Programming is introduced in Sec.~\ref{Sec8}. Section~\ref{Sec9} presents the average policy evaluation strategy via the Monte Carlo Methods. In Sec.~\ref{Sec10}, we examine the bootstrapping strategy for policy evaluation via temporal difference methods. Section~\ref{Sec11} and ~\ref{xii} introduces direct policy optimization techniques, including policy gradient and actor-critic algorithms, which are well-suited for continuous action spaces. Then, Sec.~\ref{xiv} and ~\ref{apprl} outlines how these reinforcement learning methods enable efficient, high-fidelity manipulation of quantum states, implementing quantum control. After a detailed discussion of classical reinforcement learning methods, in Sec.~\ref{QC-QML} we introduce the emerging field of quantum reinforcement learning. Here, we first provide a brief overview of the fundamentals of quantum computing, followed by a discussion of how reinforcement learning algorithms can be implemented using variational quantum circuits to realize quantum reinforcement learning.


Finally, after concluding remarks in Sec.~\ref{SecConclusion}, the appendices provide detailed discussions on the Bellman optimality condition, iterative methods, iterative policy evaluation, value iteration, as well as both first-visit and every-visit monte carlo methods.

\section{Understanding RL}\label{Sec2}

The term``reinforcement'' in RL refers to the process of encouraging or strengthening a behavior or action. For example, in a park (environment) filled with many games, giving a treat (reward) to a dog (agent) for playing specific type of games (goal) is a form of positive reinforcement. This leads the dog to learn and prefer those specific games in the park.  Over time, the dog learns to associate with these reward generating games and adjusts its behavior to play only those games that maximize its treats. With this illustration in mind, we are now in a position to define RL. Reinforcement learning is a type of machine learning where an agent interacts with an environment to achieve a goal by taking actions and receiving feedback in the form of rewards or penalties\cite{depietro2022mimo}. The agent's goal is to maximize the cumulative reward over time by learning which actions lead to the most favorable outcomes. 

Let us delve into more examples. In the field of autonomous driving, a car, acting as an agent, interacts with its road environment and receives rewards or penalties based on its actions. For example, it earns a reward for successfully avoiding a collision, while it faces a penalty for failing to stop at a red light. Over time, the car learns the best driving behaviors/action for different situations and becomes a successful self-driving car. A successful agent is the one who excels at decision making. Thus, RL helps us in producing good decision-making agents by enabling them to learn from their interactions with the environment. Note that the agent (self-driving car) learns through trials and errors. However, there is a caution. If the car is not made to interact with an environment that is complicated enough to mimic real-life traffic or road scenarios, the RL model may not generalize well when deployed in the real world. 

Another example can be taken from the field of suppy-chain management. In supply chain management, through RL we can train an agent that can optimize inventory management and demand forecasting. Here, the agent (the management system) interacts with its environment (market demand, inventory levels, production rates, and transportation costs) to perform  a suitable action (the number of items to order on a certain day) in order to balance costs, inventory, and service quality.

Note that RL is different from supervised learning in artificial intelligence. The core difference is, in supervised learning, the model learns from a labeled dataset, where the correct input-output pairs are known, whereas, in RL the model (agent) learns by interacting with an environment and receiving feedback in the form of rewards or penalties based on its actions.

\section{Essential Preliminaries}\label{Sec2.1}
To make this tutorial self-contained, we briefly describe the minimal probabilistic background, random variables and expectations. 

\textbf{Probability Theory}: Probability quantifies the likelihood of an event occurring and is constrained to the interval $[0,1]$, where 0 represents an impossible event and 1 represents certainty. For a fair coin toss, we have $\mathbb{P}(\text{Head}) = \mathbb{P}(\text{Tail}) = \frac{1}{2}$. Similarly, for a fair six-sided die, each outcome has equal probability $\mathbb{P}(\text{face } i) = \frac{1}{6}$ for $i \in \{1,2,3,4,5,6\}$.

When two events $A$ and $B$ are independent, meaning the occurrence of one does not influence the probability of the other, their joint probability follows the multiplication rule:
\begin{equation}
\mathbb{P}(A \cap B) = \mathbb{P}(A) \cdot \mathbb{P}(B)
\end{equation}
Consider a game where Alice must obtain heads on two consecutive coin tosses to win. Since coin tosses are independent events:
\begin{equation*}
\mathbb{P}(\text{Head on toss 1 and Head on toss 2}) = \frac{1}{2} \times \frac{1}{2} = \frac{1}{4}
\end{equation*}

This principle extends naturally to multiple independent events and forms the basis for many statistical models.

When events $A$ and $B$ are mutually exclusive (cannot occur simultaneously), their union probability follows the addition rule:
\begin{equation}
\mathbb{P}(A \cup B) = \mathbb{P}(A) + \mathbb{P}(B)
\end{equation}

In another game, Alice tosses a single fair six–sided die and wins if the outcome is either $1$ or $6$. 
The events “roll a $1$” and “roll a $6$” are mutually exclusive, so by the addition rule,
\begin{equation*}
\mathbb{P}(1 \text{ or } 6)
= \mathbb{P}(1) + \mathbb{P}(6)
= \frac{1}{6} + \frac{1}{6}
= \frac{1}{3}.
\end{equation*}

\textbf{Conditional Probability}: Conditional probability captures how our assessment of an event's likelihood changes when we acquire additional information. This concept is fundamental to Bayesian statistics, causal inference, and machine learning algorithms. The conditional probability of event $A$ given event $B$ is defined as:
\begin{equation}
\mathbb{P}(A \mid B) = \frac{\mathbb{P}(A \cap B)}{\mathbb{P}(B)} \quad \text{when } \mathbb{P}(B) > 0
\end{equation}
Note that, in $\mathbb{P}(A \mid B)$, the vertical bar “$\mid$” means “given,” i.e., the probability of $A$ conditioned on the event $B$.

Let us understand conditional probability through an example. Suppose in the same game of tossing two coins, Alice receives inside information that the first coin always produces heads irrespective of how it is tossed. In such a scenario, Alice's chances of winning are dramatically improved. Let us examine how conditional probability helps us quantify this advantage.

Given Alice's privileged information about the biased nature of the first coin, she knows with certainty that any outcome must begin with a head. This knowledge fundamentally transforms her probability calculations. The sample space, which originally contained four equally likely outcomes $\{HH, HT, TH, TT\}$, now reduces to just two possible outcomes: $\{HH, HT\}$. Since the second coin remains fair and independent, both remaining outcomes are equally probable.

If Alice's objective is to achieve two heads (both coins showing heads), her probability of success becomes:
\begin{equation*}
\mathbb{P}(\text{both heads}\mid\text{first coin always heads})=\frac{1}{2}.
\end{equation*}

If the first coin is guaranteed to show heads, then $\mathbb{P}(\text{first coin H})=1$ and the joint event “both heads” is equivalent to “second coin is heads.” Hence
\begin{align*}
\mathbb{P}(\text{both H}\mid\text{first coin always H})
&= \frac{\mathbb{P}(\text{both H}\cap\text{first coin H})}{\mathbb{P}(\text{first coin H})} \\
&= \frac{\mathbb{P}(\text{second coin H})}{1} \\
&= \frac{1}{2}.
\end{align*}

\textbf{Random variables}: Consider rolling a fair six-sided die. The die has 6 numbers on its faces: $\{1, 2, 3, 4, 5, 6\}$. When we roll it, a single number appears on the top face. This number changes each time we roll the die again. To capture this changing value mathematically, we define a variable $X$ and call it a \textit{random variable} that represents the number showing on top after any single roll.

The random variable $X$ can take any of the six possible values. Since the die is fair, each outcome is equally likely, so the probability of getting any specific number is the same. For example, $\mathbb{P}(X = 6) = \frac{1}{6}$, meaning there's a one-in-six chance of rolling a 6. More generally, we write $\mathbb{P}(X = k) = \frac{1}{6}$ for $k = 1, 2, 3, 4, 5, 6$. This collection of probabilities is called the probability mass function (pmf), which tells us how likely each possible outcome is.

We can also ask questions like "What is the probability of rolling 3 or less?" This leads to the cumulative distribution function (cdf), defined as $F_X(x) = \mathbb{P}(X \leq x)$. For our die example, $F_X(3) = \mathbb{P}(X \leq 3) = \mathbb{P}(X = 1) + \mathbb{P}(X = 2) + \mathbb{P}(X = 3) = \frac{3}{6} = \frac{1}{2}$.

To summarize the "typical" value we expect from rolling the die, we compute the expectation (or mean):
\begin{equation*}
\mathbb{E}[X] = \sum_{k=1}^{6} k \cdot \mathbb{P}(X = k) = 1 \cdot \frac{1}{6} + 2 \cdot \frac{1}{6} + \cdots + 6 \cdot \frac{1}{6} = \frac{21}{6} = 3.5.
\end{equation*}

\textbf{Conditional Expectation}:  Conditional expectation extends the concept of averaging to scenarios where we possess additional information. When we know that event $B$ has occurred (with $\mathbb{P}(B)>0$), the conditional expectation of a discrete random variable $X$ given $B$ computes the average using only those outcomes consistent with the observed information:
\begin{equation}
\begin{split}
\mathbb{E}[X \mid B]
&= \sum_{x} x\,\mathbb{P}(X=x \mid B)\\
&= \frac{1}{\mathbb{P}(B)} \sum_{x} x\,\mathbb{P}(X=x,\,B).
\end{split}
\label{eq:condexp-event}
\end{equation}

Returning to our die example where $X\in\{1,\dots,6\}$ represents a fair roll, suppose we learn that "the roll is at least 4." This information restricts our attention to outcomes $\{4,5,6\}$, each equally likely within this subset. The conditional expectation becomes:
\begin{equation}
\mathbb{E}[X \mid X\ge 4]
=
\frac{4\cdot \tfrac{1}{6} + 5\cdot \tfrac{1}{6} + 6\cdot \tfrac{1}{6}}{\tfrac{3}{6}}
=
\frac{4+5+6}{3}
=
5.
\end{equation}

With these essential mathematical foundations now established, one now has the theoretical toolkit necessary for understanding reinforcement learning.

\section{Terminology Alert: Part I}\label{Sec3}

In RL, we need to understand several key terms beyond the basics (agent, action, environment, goals, and rewards). These include policies, transition probabilities, state-value functions, action-value functions, episodes, trajectories, and discount factors. Let's explore each concept using a simple example.

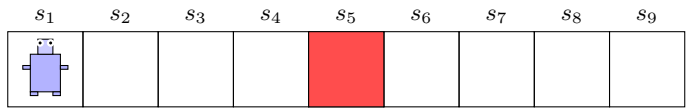
\begin{figure}
    \centering
    \begin{tikzpicture}
        \foreach \x in {1, 2, ..., 9} {
            \draw (\x, 0) rectangle (\x + 1, 1); 
            \node at (\x + 0.5, 1.2) {$s_\x$}; 
        }

        \draw[fill=blue!30] (1.3, 0.2) rectangle (1.7, 0.7); 
        \draw[fill=blue!20] (1.4, 0.7) rectangle (1.6, 0.9); 

        \fill[white] (1.43, 0.85) circle (0.05); 
        \fill[black] (1.43, 0.85) circle (0.02); 
        \fill[white] (1.57, 0.85) circle (0.05); 
        \fill[black] (1.57, 0.85) circle (0.02); 

        \draw[fill=blue!30] (1.3, 0.5) rectangle (1.2, 0.55); 
        \draw[fill=blue!30] (1.7, 0.5) rectangle (1.8, 0.55); 

        \draw[fill=blue!30] (1.3, 0.2) rectangle (1.4, 0.15); 
        \draw[fill=blue!30] (1.6, 0.2) rectangle (1.7, 0.15); 

        \draw[fill=red!70] (5, 0) rectangle (6, 1); 
    \end{tikzpicture}
    \caption{A 1D grid with a robot in cell \(s_1\) and a wall in cell \(s_5\).}
    \label{fig:grid_with_robot_and_wall}
\end{figure}

Consider a 1D grid with 9 cells (see Fig.~\ref{fig:grid_with_robot_and_wall}). Each cell in the grid is a state, hence there are 9 such states for the robot to be in. Let  $\mathcal{S} = (s_{1}, s_{2}, .., s_{9})$ be the state set. Here, the \textbf{agent} is the \textbf{robot} and the \textbf{1D grid} is the \textbf{environment} with which the agent can interact. There is no periodic boundary condition, so the agent cannot move from $s_1$ to $s_9$ and vice versa.  Further, the state $s_5$ is defined to be the terminal state, i.e., the episode terminates, and no further transitions are allowed.

Now that we have created an agent and the environment, it is time to introduce actions, goal, and reward system to achieve the goal. In the 1D grid environment, the agent in any state is allowed to take any of the two \textbf{actions} -- $\mathcal{A}$ = (\texttt{"right"}, \texttt{"left"}). Note that  if the agent takes action \texttt{"left"} by being in state $s_{1}$ or action \texttt{"right"} in state $s_{9}$, the state is the same as the current state. The goal of the agent is to reach the terminal state $s_{5}$ following a policy (defined below). To determine which actions are preferable, we introduce a reward function that assigns a numerical reward or penalty to each action taken and its resulting outcome. Our objective is to encourage the agent to reach a terminal state as quickly as possible. Therefore, we can assign a negative reward (penalty) for actions that move the agent away from the goal or increase the time to reach it, and a positive reward for successfully completing the task. Additionally, we can define a cumulative reward over an entire trajectory (episode), and train the agent to maximize the total expected return--that is, the sum of rewards collected throughout the episode. Hence, the \textbf{reward system} is chosen such that the agent gets a reward of -1 for each step it takes toward any regular state, but receives +5 when it reaches the terminal (goal) state.

\textbf{Policy}: Now let us design a \textbf{policy} for the agent to achieve this goal. We will start with the formal mathematical definition, then illustrate it with a concrete example.

A policy $\pi(a|s)$ is a mapping from states to probabilities over actions, 
\begin{equation}
\pi: \mathcal{S} \rightarrow \mathcal{P}(\mathcal{A})
\end{equation}
where $\mathcal{S} $ is the set of all possible states, $ \mathcal{A} $ is the set of all possible actions,
and $ \mathcal{P}(\mathcal{A}) $ is the set of probability distributions over $ \mathcal{A} $. Thus, for any state $ s $,  the policy $ \pi(a|s) $ gives the probability of choosing action $ a $ in that state. The policy essentially defines a probability distribution over actions for each state, satisfying
\begin{equation}
    \sum_{a \in \mathcal{A}} \pi(a|s) = 1 \quad \text{for all } s \in \mathcal{S}.
\end{equation}

We start with a deterministic policy specified by Table~\ref{policy_mapping}. For the toy example of Fig.~\ref{fig:grid_with_robot_and_wall}, it should be obvious that such a policy  is not optimal. However, we will show below how to  improve it iteratively.
\begin{table}
    \centering
    \begin{tabular}{|c|c|c|}
        \hline
        \textbf{State} & $ \mathcal{P}(right) $ & $ \mathcal{P}(left) $ \\
        \hline
        $s_{1}$ & 1 & 0 \\
        $s_{2}$ & 1 & 0 \\
        $s_{3}$ & 1 & 0 \\
        $s_{4}$ & 1 & 0 \\
        $s_{5}$ & 0 & 0 \\ 
        $s_{6}$ & 1 & 0 \\
        $s_{7}$ & 1 & 0 \\
        $s_{8}$ & 1 & 0 \\
        $s_{9}$ & 1 & 0 \\
        \hline
    \end{tabular}
    \caption{Policy $\pi$ mapping for the 1D grid example.}
    \label{policy_mapping}
\end{table}

The information in Table~\ref{policy_mapping} can be simply written in the form of a Python dictionary as follows, 

\begin{lstlisting}
policy = {
    "s1": {"right": 1, "left": 0},                
    "s2": {"right": 1, "left": 0},   
    "s3": {"right": 1, "left": 0},   
    "s4": {"right": 1, "left": 0},                           
    "s5": {},  # Terminal state
    "s6": {"right": 1, "left": 0},  
    "s7": {"right": 1, "left": 0},   
    "s8": {"right": 1, "left": 0},  
    "s9": {"right": 1, "left": 0}
}
\end{lstlisting}

Note that the policy was implemented in Python using the built-in dictionary~\footnote{Please consult the official documentation \url{https://docs.python.org/3/tutorial/datastructures.html#dictionaries} for details about Python dictionaries.} data type. A dictionary stores data as key-value pairs, where each key maps to its corresponding value. For instance, \lstinline{python}{"s1"} is a key that maps to a nested dictionary \lstinline{python}{{"right": 1, "left": 0}}. The key advantage of dictionaries is their efficient value retrieval: given a key, the corresponding value can be extracted in constant time, regardless of the dictionary's size! For example, the command \lstinline{python}{policy["s1"]} returns the associated value with $O(1)$ time complexity. This efficiency is achieved through hash-based storage, which does not preserve insertion order~\footnote{Note there is a separate data type in Python for ordered dictionaries \url{https://docs.python.org/3/library/collections.html#ordereddict-objects}}.

 For the policy in Table \ref{policy_mapping}, $\pi(right|s_{2}) = 1$ and $\pi(left|s_{2}) = 0$ and therefore we have
 \begin{equation}
     \sum_{a \in \mathcal{A}} \pi(a|s_{2})= \pi(right|s_{2}) + \pi(left|s_{2})  = 1.
 \end{equation}

Note that the policy serves as a guiding framework for the agent, indicating which action to take when it is in a specific state. In our case, the policy (see Table \ref{policy_mapping}) is  deterministic because it allows the agent to take the action  $\texttt{"right"}$ for any state it is in. However, at this point, a natural question arises: What is the probability that this action will lead the agent to end up in state $s_2$? In RL, uncertainty is inherent in the agent's interactions with its environment. For instance, when the agent is in state $s_1$ and decides to take the action $\texttt{"right"}$, we cannot definitively conclude that it will end up in state $s_2$. This uncertainty arises because we have not previously specified the potential outcomes associated with the actions available to the agent in each state. This question leads us to the concept of \textbf{state-transition probabilities}, denoted as $p(s'| s, a)$.
 
The state-transition probability $p(s'| s, a)$ is a three-argument function:
\begin{equation}
   \mathcal{P}: \mathcal{S} \times\mathcal{S} \times\mathcal{A} \rightarrow [0,1].
\end{equation}
This represents the probability that the agent transitions to state + $s'$ given current state $s$ and action $a$. In other words, for any state-action pair $(s,a)$, $p(s'|s, a)$ gives the probability of moving to the next state $s'$. As a probability distribution, it must satisfy
\begin{equation}
     \sum_{s' \in \mathcal{S}} p(s'|s, a) =  1.
     \label{st-equation}
\end{equation}

To simplify our analysis, we use deterministic state transitions where probabilities are either 0 or 1. When an agent takes an action from a given state, it transitions to a single predetermined next state with probability 1, while all other transitions have probability 0.

Consider the example of state transition probabilities in Table~\ref{state_transition_probabilities}.
\begin{table}
    \centering
    \begin{tabular}{|c|c|c|c|c|}
        \hline
        \textbf{State} & \textbf{Action} & \textbf{Next State ($s'$)} & \textbf{Probability} \\
        \hline
        $s_{1}$ & right & $s_{2}$ & 1 \\
        \hline
        $s_{1}$ & left & $s_1$ & 1 \\
        \hline
        $s_{2}$ & right & $s_{3}$ & 1 \\
        \hline
        $s_{2}$ & left & $s_{1}$ & 1 \\
        \hline
        $s_{3}$ & right & $s_{4}$ & 1 \\
        \hline
        $s_{3}$ & left & $s_{2}$ & 1 \\
        \hline
        $s_{4}$ & right & $s_{5}$ & 1 \\ 
        \hline
        $s_{4}$ & left & $s_{3}$ & 1 \\
        \hline
        $s_{5}$ & right & $s_{5}$ & 1 \\
        \hline
        $s_{5}$ & left & $s_{5}$ & 1 \\
        \hline
        $s_{6}$ & right & $s_{7}$ & 1 \\
        \hline
        $s_{6}$ & left & $s_{5}$ & 1 \\
        \hline
        $s_{7}$ & right & $s_{8}$ & 1 \\
        \hline
        $s_{7}$ & left & $s_{6}$ & 1 \\
        \hline
        $s_{8}$ & right & $s_{9}$ & 1 \\
        \hline
        $s_{8}$ & left & $s_{7}$ & 1 \\
        \hline
        $s_{9}$ & right & $s_{9}$ & 1 \\ 
        \hline
        $s_{9}$ & left & $s_{8}$ & 1 \\ 
        \hline
    \end{tabular}
    \caption{Intuitive State Transition Probabilities $P(s'|s,a)$ indicating next states and their probabilities. }
    \label{state_transition_probabilities}
\end{table}
Due to the deterministic nature, the transition probability can be represented by a Python dictionary.
\begin{lstlisting}
   transitions = {
    "s1": {"right": "s2", "left": "s1"},          
    "s2": {"right": "s3", "left": "s1"}, 
    "s3": {"right": "s4", "left": "s2"},
    "s4": {"right": "s5", "left": "s3"},
    "s5": {},  # Terminal state 
    "s6": {"right": "s7", "left": "s5"}, 
    "s7": {"right": "s8", "left": "s6"},
    "s8": {"right": "s9", "left": "s7"},
    "s9": {"right": "s9", "left": "s8"}
} 
\end{lstlisting}

It is natural to wonder what happens when we have non-deterministic transitions and what might cause that. For example, if $p(s_{2} | s_{1}, right) = 0.7$ and $p(s_{4} | s_{1}, right) = 0.3$, then taking action {``right''} from  state $s_{1}$ leads  to state $s_{2}$ with 70\% probability and to state $s_{4}$ with 30\% probability.

This uncertainty might come from a ``mysterious force'' in the environment, which adds randomness to what happens next. Even though the agent can choose its actions based on its current state and policy, the outcome is not guaranteed. So, while the chance of ending up in state $s_{4}$ is lower at 30\%, it is still possible, showing that decision-making can be complicated in uncertain situations. In real-life scenarios, RL environments are often stochastic due to various uncertainties and complexities inherent in the real world. For example, an RL environment where human decisions are involved and given the unpredictability of human behavior and decision, the environment can itself become stochastic. In reinforcement learning, the agent interacts with the environment and receives rewards, which are numerical values indicating how good or bad a taken action was. These rewards guide learning by encouraging the agent to choose actions that maximize long-term benefit.

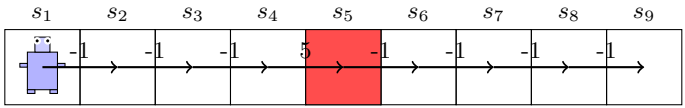
\begin{figure}
    \centering
    \begin{tikzpicture}
        \foreach \x in {1, 2, ..., 9} {
            \draw (\x, 0) rectangle (\x + 1, 1); 
            \node at (\x + 0.5, 1.2) {$s_\x$}; 
        }

        \draw[fill=blue!30] (1.3, 0.2) rectangle (1.7, 0.7); 
        \draw[fill=blue!20] (1.4, 0.7) rectangle (1.6, 0.9); 

        \fill[white] (1.43, 0.85) circle (0.05); 
        \fill[black] (1.43, 0.85) circle (0.02); 
        \fill[white] (1.57, 0.85) circle (0.05); 
        \fill[black] (1.57, 0.85) circle (0.02); 

        \draw[fill=blue!30] (1.3, 0.5) rectangle (1.2, 0.55); 
        \draw[fill=blue!30] (1.7, 0.5) rectangle (1.8, 0.55); 

        \draw[fill=blue!30] (1.3, 0.2) rectangle (1.4, 0.15); 
        \draw[fill=blue!30] (1.6, 0.2) rectangle (1.7, 0.15); 

        \draw[fill=red!70] (5, 0) rectangle (6, 1); 
        
        \foreach \x in {1, 2, ..., 8} {
            \draw[->, thick] (\x + 0.5, 0.5) -- (\x + 1.5, 0.5) 
            node[midway, above] {\ifnum\x=4 5 \else -1 \fi}; 
        }
    \end{tikzpicture}
    \caption{An improved version of Figure~\ref{fig:grid_with_robot_and_wall} with action and rewards explicitly shown.}
    \label{fig:rewardfig}
\end{figure}

Before going further, we graphically represent the introduced concepts in  Fig.~\ref{fig:rewardfig}. 

\textbf{Episode and Trajectory}: Let us assume that our agent in the 1D grid world follows the policy $\pi$. The goal is to reach the terminal state $s_{5}$. If the agent begins from the current state $s_{1}$ as shown in Figure~\ref{fig:rewardfig} and follows the policy $\pi$, it reaches the terminal state $s_{5}$ in a finite number of steps. In this context, an episode begins when the agent starts in state $s_{1}$. The episode progresses as the agent follows the actions dictated by policy $\pi$ and  moves through various states in the grid. The episode continues until the agent successfully reaches the terminal state $s_{5}$. Throughout the episode, the agent collects rewards.

The \textbf{trajectory}, on the other hand, demonstrates how the agent navigates through the grid world with the decision it makes at each state according to policy $\pi$ until it reaches the terminal state $s_{5}$ where the episode ends. A \textbf{trajectory} can be described as a set that contains tuples of the form $(s_t, a_t, r_{t+1})$, where $s_t$ represents the current state of the agent at time $t$, $ a_t$ denotes the action taken by the agent in state $s_t$, $r_{t+1}$ signifies the reward received after taking action $a_t$. In our case, for the episode where the agent begins in the current state $s_{1}$ and reaches the terminal state $s_{5}$, the trajectory is (in Python list and tuples): 
\begin{lstlisting}
Trajectory = [
    ("s1", "right", -1), 
    ("s2", "right", -1),
    ("s3", "right", -1),
    ("s4", "right", 5)
]
\end{lstlisting}
To connect this with the mathematical notation, we align the Python labels 
with the time-indexed states as follows:

\[
\begin{array}{ccl}
\text{``s1''} & \;\;\longleftrightarrow\;\; & s_0 \quad (\text{initial state at } t=0) \\
\text{``s2''} & \;\;\longleftrightarrow\;\; & s_1 \\
\text{``s3''} & \;\;\longleftrightarrow\;\; & s_2 \\
\text{``s4''} & \;\;\longleftrightarrow\;\; & s_3 \\
\end{array}
\]

Thus, a trajectory is 
\[
 \left[
    (s_0, a_0, r_{1}),
    (s_1, a_1, r_{2}),
    (s_2, a_2, r_{3}),
    (s_3, a_3, r_{4})
\right]
\]

\textbf{State Value Function}: The state value function of a state $s$ under a policy $\pi$, denoted $v_{\pi}(s)$, is the expected return when starting in $s$ and following policy $\pi$ thereafter \cite{Sutton2018-bn}. Let us understand the definition in our context. Given the agent is in state $s_{1}$ and follows policy $\pi$, the return is the sum of rewards we see in the trajectory above. Note that \textbf{reward} refers to the immediate feedback received after each action, while \textbf{return} $G_{t}$ represents the cumulative return from time $t$ onward, defined as the sum of future rewards received by the agent (or sum of future rewards starting from the current state until the end of the episode in an episodic task)
\begin{equation}
    G_{t} = r_{t+1} + r_{t+2} + ... + r_{T}.
\end{equation}
where $T$ is the final time step that leads the agent to the terminal state.
In our case, 
\begin{align*}
    \text{Return} &= \text{Sum of future rewards} \\  &= (-1) + (-1) + (-1) + (+5) = +2.
\end{align*}
 Since the trajectory is deterministic (there is no variability in the trajectory across different episodes starting from the current state, say $s_{1}$), in every run, the sum of reward/return is the same. Therefore, the value of a state in this context can be directly computed as the total return from that state to the end of the episode, and the expectation becomes unnecessary. Thus, the value of a function for the deterministic case as ours can be mathematically defined as
 \begin{equation}
     v_{\pi}(s) = G_{t} |_{S_{t} = s}
 \end{equation}
where $S_{t} = s$ indicates the initial state of the agent/state from which the agent begins following the policy.

Let's examine which state is more beneficial for the agent to be in. A quick calculation reveals that state $ s_4 $ has a value of +5, which is higher than any other state. This value is expected to be elevated since it is closer to the terminal state.

Expectation is essential when the environment is stochastic, which means there is randomness in the transitions or rewards, or both. This randomness can lead to various possible outcomes for the agent's trajectory, even if it starts from the same state and follows the same policy every time. In such a case, the value of a state can be defined mathematically (take $\gamma = 1$) as:
\begin{align}
v_\pi(s) = \mathbb{E}_\pi \left[ \sum_{k=0}^{T-t-1} r_{t+k+1} \mid S_t = s \right]
\end{align}

The state-value function under a policy \( \pi \), denoted \( v_\pi(s) \), represents the expected return when starting in state \( s \) and following policy \( \pi \) thereafter. It is formally defined as:
\begin{align}
    v_\pi(s) \doteq \mathbb{E}_\pi\left[G_t \mid S_t = s\right] 
    = \mathbb{E}_\pi\left[\sum_{k=0}^{T-t-1} \gamma^k r_{t+k+1} \,\middle|\, S_t = s\right],
    \label{value_function_stochastic}
\end{align}
where:
\( \gamma \in [0, 1] \) is the discount factor,  \( r_{t+k+1} \) is the reward received at time step \( t+k+1 \),
 \( G_t = \sum_{k=0}^{T-t-1} \gamma^k r_{t+k+1} \) is the return from time \( t \),
 and \( \mathbb{E}_\pi[\cdot] \) denotes the average / expectation over all possible trajectories generated by following policy \( \pi \), i.e., it accounts for both the stochasticity of the environment and the action selection governed by \( \pi \).


\textbf{Discount Factor ($\gamma$):} Up until now, we discussed agent-environment interaction that can be conceptualized in the form of episodes. In such cases, there is a clear starting point and an ending point (when the agent hits the terminal state). However, there are cases when agent-environment interaction can go on continuously without a well-defined ending point. In cases like these the idea of episodes does not apply. An example of this ongoing, non-episodic interaction is temperature control in an industrial environment, where the system constantly adjusts heating or cooling levels to maintain optimal conditions. Here, there is no natural ``end'' to the process—the controller operates continuously without a reset, adjusting based on a feedback loop that runs indefinitely.

In cases of continuous interaction between the agent and environment, defining the return $G_{t}$ can be challenging because it may grow indefinitely. To keep the return manageable, we use a discount factor, which ensures that $G_{t}$ stays finite by reducing the impact of future rewards over time. In such cases, we have,

\begin{equation}
    G_t \doteq r_{t+1}+\gamma r_{t+2}+\gamma^2 r_{t+3}+\cdots=\sum_{k=0}^{\infty} \gamma^k r_{t+k+1},
    \label{Discount_factor}
\end{equation}
where an extra $\gamma$ factor is introduced such that for $\gamma < 1$, the above sum stays finite.

\section{Terminology Alert: Part II}\label{Sec4}

In the above section, we explained basic terminology that \textbf{one} encounters in studying RL. In this section, we will explain other terminologies, such as state-reward transition probability and action-value functions.

\textbf{State-Reward Transition Probability}: In the previous section, we understood state-transition probability $p(s'|s,a)$ which gives the probability for an agent to reach state $ s' $ given it is in state $s$ and takes action $a$ being in the current state. Similarly, state-reward transition probability $p(s',r|s,a)$ is the probability for the agent to reach state $s'$ and simultaneously achieve reward $r$ given  it is in state $s$ and takes action $a$ being in the current state. The relation between both quantities is as follows:
\begin{equation}
    \sum_{r} p(s',r|s, a) =  p(s'|s,a)
     \label{transition-relations-equation}
\end{equation}

In our example of 1D grid, $p(s_{2},5|s_{1},right) = 0$ but $p(s_{2}|s_{1},right) = 1$. This is because, in the former case, a transition from $s_{1}$ to $s_{2}$ does take place when the action taken is $right$ but the reward is not 5 instead it is -1. In the latter case, it is straightforward to see that it is indeed true. Note that the function $ p(s' | s, a) $ shows the likelihood of reaching a new state but offers no information about rewards, which the agent seeks to maximize. Without reward information, the agent may end up in desirable states without knowing which actions provide the most benefit. In contrast, $ p(s', r | s, a) $ gives the agent a clearer picture by considering both next states and their associated rewards, enabling better decision-making.

The environment's uncertainty is encapsulated in the state-reward transition probability $p(s', r \mid s, a)$. In a deterministic setting, this probability is either 0 or 1, indicating a certain outcome. In a non-deterministic environment, it includes probability values and becomes non-trivial.

\textbf{Action-Value Function}: In the previous section, we explained value function of a state. The action-value function, denoted $q_{\pi}(s, a)$, is the expected return when starting in state $s$, taking a specific action $a$, and then following the policy $\pi$ from the resulting state. Mathematically, the action-value function is

\begin{align}
    q_\pi(s,a) &= \mathbb{E}_\pi\left[G_t \mid S_t=s,A_t= a\right] \\  &=  \mathbb{E}_\pi\left[\sum_{k= 0}^{T-t-1} \gamma^k r_{t+k+1} \middle|\ S_t=s, A_t = a\right]. 
\end{align}

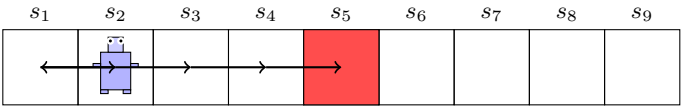
\begin{figure}
    \centering
    \begin{tikzpicture}
        \foreach \x in {1, 2, ..., 9} {
            \draw (\x, 0) rectangle (\x + 1, 1); 
            \node at (\x + 0.5, 1.2) {$s_\x$}; 
        }

        \draw[fill=blue!30] (2.3, 0.2) rectangle (2.7, 0.7); 
        \draw[fill=blue!20] (2.4, 0.7) rectangle (2.6, 0.9); 

        \fill[white] (2.43, 0.85) circle (0.05); 
        \fill[black] (2.43, 0.85) circle (0.02); 
        \fill[white] (2.57, 0.85) circle (0.05); 
        \fill[black] (2.57, 0.85) circle (0.02); 

        \draw[fill=blue!30] (2.3, 0.5) rectangle (2.2, 0.55); 
        \draw[fill=blue!30] (2.7, 0.5) rectangle (2.8, 0.55); 

        \draw[fill=blue!30] (2.3, 0.2) rectangle (2.4, 0.15); 
        \draw[fill=blue!30] (2.6, 0.2) rectangle (2.7, 0.15); 

        \draw[fill=red!70] (5, 0) rectangle (6, 1); 

        \draw[->, thick] (2.5, 0.5) -- (1.5, 0.5); 
        \draw[->, thick] (2.5, 0.5) -- (3.5, 0.5); 
        \foreach \x in {1, 3, 4} {
            \draw[->, thick] (\x + 0.5, 0.5) -- (\x + 1.5, 0.5); 
        }
    \end{tikzpicture}
    \caption{Another version of Fig.~\ref{fig:grid_with_robot_and_wall} with robot in cell 2 taking first action to left and then following policy $\pi$}
    \label{actionvaluerewardfig}
\end{figure}

Let us understand this through an example. Note that in the deterministic case, expectation is not necessary (as explained before). In our 1D grid example, assume that the agent is allowed to take action $\texttt{"left"}$. Now if the agent being is state $s_{2}$ takes the first action  $\texttt{"left"}$ and then follow the policy (see Fig.~\ref{actionvaluerewardfig}), then in such case the trajectory is

\begin{lstlisting}
Trajectory = [
    ("s2", "left", -1), 
    ("s1", "right", -1),
    ("s2", "right", -1),
    ("s3", "right", -1),
    ("s4", "right", 5)
]
\end{lstlisting}

Note that since there is a natural temporal order in a trajectory, we implemented \lstinline{python}{Trajectory} as a Python list~\footnote{\url{https://docs.python.org/3/tutorial/datastructures.html#more-on-lists}} of tuples~\footnote{\url{https://docs.python.org/3/tutorial/datastructures.html#tuples-and-sequences}}. Lists allow efficient extraction of elements (i.e., within the $O(1)$ time) by their ordinal index. However, Python indexing starts from $0$. For example, \lstinline{python}{Trajectory[0]} returns \lstinline{python}{("s1", "right", -1)}, while \lstinline{python}{Trajectory[3]} returns \lstinline{python}{("s4", "right", 5)}.

From the trajectory we have a return \(G_{t} = 1\). Therefore,
\[
q_{\pi}(s_{1}, \text{left}) = 1,
\]
because the only observed return following action \(\text{left}\) in state \(s_{1}\) is \(1\).

There exists a direct relationship between the two value functions. Mathematically, they are related as follows:
\begin{equation}
v_{\pi}(s) = \sum_{a} \pi(a | s) q_{\pi}(s, a)
\end{equation}

A crucial difference between state-value function and action-value function is that the former is useful for evaluating the value of states, regardless of the action, as it averages over all possible actions according to the policy $\pi$ and the latter is helpful when an agent needs to decide between different actions in a given state since it assesses each action’s potential return individually \cite{tutorial1}.


\section{Objective of RL Part I: Setting the Stage}\label{Sec5}

The primary goal of reinforcement learning (RL) is to enable an agent to learn effective decision-making. This is accomplished by having the agent interact with the environment and use the reward system to refine its policies until it reaches an optimal policy, which ultimately enhances its decision-making abilities. 

\begin{figure}
    \centering
    \begin{tikzpicture}
        \foreach \x in {1, 2, ..., 9} {
            \draw (\x, 0) rectangle (\x + 1, 1); 
            \node at (\x + 0.5, 1.2) {$s_\x$}; 
        }

        \draw[fill=blue!30] (1.3, 0.2) rectangle (1.7, 0.7); 
        \draw[fill=blue!20] (1.4, 0.7) rectangle (1.6, 0.9); 

        \fill[white] (1.43, 0.85) circle (0.05); 
        \fill[black] (1.43, 0.85) circle (0.02); 
        \fill[white] (1.57, 0.85) circle (0.05); 
        \fill[black] (1.57, 0.85) circle (0.02); 

        \draw[fill=blue!30] (1.3, 0.5) rectangle (1.2, 0.55); 
        \draw[fill=blue!30] (1.7, 0.5) rectangle (1.8, 0.55); 

        \draw[fill=blue!30] (1.3, 0.2) rectangle (1.4, 0.15); 
        \draw[fill=blue!30] (1.6, 0.2) rectangle (1.7, 0.15); 

        \draw[fill=red!70] (5, 0) rectangle (6, 1); 
        
        \foreach \x in {1, 2, 3} {
            \draw[<-, thick] (\x + 1.5, 0.5) -- (\x + 0.5, 0.5) 
            node[midway, above] {-1}; 
        }
        \foreach \x in {4, 5, 6, 7, 8} {
            \ifnum\x=4
                \draw[<-, thick] (\x + 1.4, 0.5) -- (\x + 0.4, 0.5) 
                node[midway, above] {5}; 
            \else
                \ifnum\x=5
                    \draw[<-, thick] (\x + 0.6, 0.5) -- (\x + 1.6, 0.5) 
                    node[midway, above] {5}; 
                \else
                    \draw[<-, thick] (\x + 0.5, 0.5) -- (\x + 1.5, 0.5) 
                    node[midway, above] {-1}; 
                \fi
            \fi
        }

    \end{tikzpicture}
    \caption{Updated version of Fig.~\ref{fig:rewardfig} with improved policy}
    \label{fig:improvedpolicy}
\end{figure}
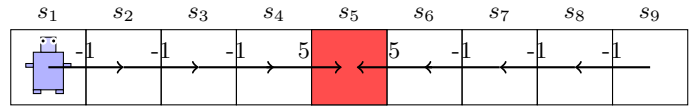

Let us illustrate policy improvement with an example based on our 1D grid scenario. The original policy allows the agent to reach the terminal state $s_{5}$ when the agent's current state is any position to the left of $s_{5}$. However, if the agent finds itself in any state to the right of the terminal state, the current policy does not provide a way to reach the terminal state. Consequently, this policy is not effective and requires improvement. A straightforward enhancement to the policy would be to ensure that when the agent is in any state to the right of the terminal state, the policy allows only the \texttt{"left"} action, instead of permitting the \texttt{"right"} action. Thus the improved policy (see also Fig.~\ref{fig:improvedpolicy}) is
\begin{lstlisting}
improved_policy = {
    "s1": {'right': 1, 'left': 0},                
    "s2": {'right': 1, 'left': 0},   
    "s3": {'right': 1, 'left': 0},   
    "s4": {'right': 1, 'left': 0},                           
    "s5": {},  # Terminal state
    "s6": {'right': 0, 'left': 1},  
    "s7": {'right': 0, 'left': 1},   
    "s8": {'right': 0, 'left': 1},  
    "s9": {'right': 0, 'left': 1}
}
\end{lstlisting}

At this point, a natural question to ask is: How can we improve the policy? The above example was trivial; hence, quick intuition helped us figure out what might be a good policy. However, in more complicated cases, we need algorithms to assist us in the policy improvement process. Before we move on to algorithms, let us understand what fundamentally dictates the policy improvement process.

The value functions defined previously can be used to dictate what constitutes a better policy. A policy $ \pi $ is at least as good as policy $ \pi' $ (denoted $ \pi \succeq \pi' $) if and only if \cite{Sutton2018-bn}:
\begin{equation}
    v_{\pi}(s) \geq v_{\pi'}(s), \quad \text{for all } s \in \mathcal{S}.
\end{equation}
As the agent explores the environment, it improves the values of states (by exploring and updating the values of states based on received rewards) in order to refine the policy. As the iterative process continues, the agent eventually reaches an optimal policy, where it has maximized its expected reward from each state. Thus, an optimal policy is the policy that maximizes the value of all states. Mathematically, for an optimal policy $\pi = *$, we have
\begin{align}
    v_{*}(s) = \max _{\pi} v_{\pi}(s), \quad \text{for all } s \in \mathcal{S},
\end{align}
where $v_{*}(s)$ is the optimal state-value function that one achieves when the policy is optimal. In our case, if the improved policy described above is an optimal policy, then the value of all states $s$ is the optimal value for those states. It is easy to list down the optimal value of the states in this case (note the value of a terminal state is always zero as there is no future reward to collect from there): 

\begin{lstlisting}
state_values = {
    "v(s1)": 2,
    "v(s2)": 3,
    "v(s3)": 4,
    "v(s4)": 5,
    "v(s5)": 0,  # Terminal state
    "v(s6)": 5,
    "v(s7)": 4,
    "v(s8)": 3,
    "v(s9)": 2
}
\end{lstlisting}

It's important to recognize that optimal policies also have the same optimal action-value function, denoted $q_*$, and defined as \cite{Sutton2018-bn}:
\begin{equation}
    q_*(s, a) \doteq \max _\pi q_\pi(s, a),
\end{equation}

for all $s \in \mathcal{S}$ and $a \in \mathcal{A}(s)$, where, $\mathcal{A}(s)$ is the set of actions available in state $s$.

\section{Objective of RL Part II: Policy Evaluation and Policy Improvement}\label{objective2}

When an agent lacks complete information about its environment, it learns by interacting with it, often using a trial-and-error approach to improve its policy. In these situations, the agent may encounter multiple possible episodes that can start from the same state, say $s_{2}$. To find the value of that state, the agent needs to look at the outcomes of all these episodes. By adding up the values from each episode and averaging them, the agent can better understand the state $s_{2}$ value. In the 1D grid example, it is possible to have many episodes starting from a given state for non-trivial state-transition probabilities. For example, a non-trivial transition with probability (in Python dictionary) is as follows:
\begin{lstlisting}
   transitions = {
    "s1": {
        "right": [("s2", 0.7), ("s1", 0.3)],
        "left": [("s1", 1.0)]
    },
    "s2": {
        "right": [("s3", 0.8), ("s1", 0.2)],
        "left": [("s1", 1.0)]
    },
    "s3": {
        "right": [("s4", 0.9), ("s2", 0.1)],
        "left": [("s2", 1.0)]
    },
    "s4": {
        "right": [("s5", 1.0)],
        "left": [("s3", 1.0)]
    },
    "s5": {}, #Terminal state
    "s6": {
        "right": [("s7", 0.6), ("s5", 0.4)],
        "left": [("s5", 1.0)]
    },
    "s7": {
        "right": [("s8", 0.7), ("s6", 0.3)],
        "left": [("s6", 1.0)]
    },
    "s8": {
        "right": [("s9", 0.8), ("s7", 0.2)],
        "left": [("s7", 1.0)]
    },
    "s9": {
        "right": [("s9", 1.0)],
        "left": [("s8", 1.0)]
    }
\end{lstlisting}

This clearly states that the 1D grid environment is now more stochastic in nature. For example, given an agent is state $s_{2}$ takes action $\texttt{"right"}$, the agent has some probability ($20 \%$) that it can end up in state $s_{1}$. With this uncertainty, every episode (and its return) starting from a given state will be different.

We will now focus on an environment where complete information is available, specifically regarding the transition probabilities \( p(s', r \mid s, a) \). This means we have a thorough understanding of how the environment responds to the agent's actions, including the resulting states and associated rewards. In this case, we do not need to resort to trial and error to determine state values. Instead, we can utilize what is known as the Bellman equation to find the value of each state as shown in Appendix~\ref{bellappen}. The Bellman equation is expressed as follows: 
\begin{equation}
  v_\pi(s) = \sum_a \pi(a \mid s) \sum_{\substack{s' \in \mathcal{S} \setminus \{s\} \\ r \in \mathcal{R}}}  p(s', r \mid s, a) \left[ r + \gamma v_\pi(s') \right]
  \label{bellman}
\end{equation}
for all $s \in \mathcal{S}$, and $s' \in \mathcal{S} \text{ such that } s' \neq s$. This can be written in a mathematical language as  $\forall s \in \mathcal{S}, \quad s' \in \mathcal{S} \setminus \{s\}$.

To understand the Bellman equation Eq.~\eqref{bellman} and its significance, consider an environment where the transition probabilities $p(s', r \mid s, a)$ are known. Given a policy and assuming that the values of all states $s'$ are known (i.e., $ v_{\pi}(s')$ are known for $s' \in \mathcal{S} \setminus \{s\}$), we can solve the Bellman equation to determine the value of a particular state $s$. 

Let us illustrate this through an example. Consider a $2 \times 2$ grid with state set $\mathcal{S} = (s_{1}, s_{2}, s_{3}, s_{4})$. Further $\mathcal{S} \setminus \{s_{1}\} = ( s_{2}, s_{3}, s_{4}) $, $\mathcal{S} \setminus \{s_{2}\} = ( s_{1}, s_{3}, s_{4}) $, $\mathcal{S} \setminus \{s_{3}\} = ( s_{1}, s_{2}, s_{4}) $ and $\mathcal{S} \setminus \{s_{4}\} = ( s_{1}, s_{2}, s_{3}) $ . Let $s_{4}$ be the terminal state. The reward and action sets are $\mathcal{R} = (-1,3)$ and $\mathcal{A} = (\texttt{"right"}, \texttt{"left"},  \texttt{"up"}, \texttt{"down"})$ respectively. Further, the policy and transitions are as follows: 

\begin{lstlisting}
policy = {
    "s1": {'right': 0.5, 'down': 0.5},                
    "s2": {'left': 0.5, 'down': 0.5},   
    "s3": {'right': 0.5, 'up': 0.5},   
    "s4": {} #Terminal state
}
\end{lstlisting}
\begin{lstlisting}
   transitions = {
    "s1": {"right": "s2", "down": "s3"},          
    "s2": {"left": "s1", "down": "s4"}, 
    "s3": {"right": "s4", "up": "s1"},
    "s4": {} #Terminal state
} 
\end{lstlisting}

Intuitively, it is clear that the policy is not an optimal policy and needs improvement. However, in order to solve the Bellman equation, we will use this policy (with discount factor $\gamma = 1$). 

We will solve the Bellman equation analytically (note that the Bellman equation in practice is solved numerically). Before solving the equation, let us note the following:
In Eq.~\eqref{bellman}, for the above example, we have  $\pi(right | s_{1}) = \pi(down | s_{1}) = 0.5 $ and $\pi(left | s_{1}) = \pi(up | s_{1}) = 0 $. Similarly, for other states, these quantities can be obtained by looking at the policy given above in Python dictionary format. Likewise, for the transition probabilities, note that for state $s_{1}$, $p(s_{2}, -1 | s_{1}, right) = p(s_{3}, -1 | s_{1}, \text{down}) = 1$, while all other transitions from $s_{1}$ have zero probability. Transition probabilities for other states can be similarly obtained by referring to the dictionary format provided above.

\begin{equation}
\begin{aligned}
    v_\pi(s_{1}) &= \sum_{a \in \mathcal{A}} \pi(a \mid s_{1}) \sum_{\substack{s' \in \mathcal{S} \setminus \{s_{1}\} \\ r \in \mathcal{R}}} p(s', r \mid s_{1}, a) \left[ r + v_\pi(s') \right] \\
    &= \sum_{a \in \mathcal{A}} \pi(a \mid s_{1}) \Big\{ p(s_{2}, -1 \mid s_{1}, a) \left[ -1 + v_{\pi}(s_{2}) \right] \\
    &\quad + p(s_{3}, -1 \mid s_{1}, a) \left[ -1 + v_{\pi}(s_{3}) \right] \Big\} \\
    &= 0.5 \Big\{ 1 \left[ -1 + v_{\pi}(s_{2}) \right] + 1 \left[ -1 + v_{\pi}(s_{3}) \right] \Big\} \\
    &= 0.5 \left[ -1 + v_{\pi}(s_{2}) \right] + 0.5 \left[ -1 + v_{\pi}(s_{3}) \right] \\
    &= -1 + 0.5 v_{\pi}(s_{2}) + 0.5 v_{\pi}(s_{3}) 
    \label{eq20}
\end{aligned}
\end{equation}
Similarly, for state $s_{3}$, we have 

\begin{equation}
\begin{aligned}
v_\pi(s_3) &= \sum_{a \in \mathcal{A}} \pi(a \mid s_3) \sum_{\substack{s' \in \mathcal{S} \setminus \{s_3\} \\ r \in \mathcal{R}}} p(s', r \mid s_3, a) \left[ r + v_\pi(s') \right] \\
&= \pi(a = right \mid s_3) \left[ p(s_4, 3 \mid s_3, right) \left( 3 + v_\pi(s_4) \right) \right] \\
& \quad + \pi(a = \text{up} \mid s_3) \left[ p(s_1, -1 \mid s_3, \text{up}) \left( -1 + v_\pi(s_1) \right) \right] \\
&= 0.5 \left[ 3 + v_\pi(s_4) \right] + 0.5 \left[ -1 + v_\pi(s_1) \right] \\
&= 1.5 + 0.5 v_\pi(s_1) - 0.5 + 0.5 v_\pi(s_4)\\
&= 1 + 0.5 v_\pi(s_1)  + 0.5 v_\pi(s_4)\\
\end{aligned}
\end{equation}

Thus, Bellman equation gives us a set of four equations as follows:
\begin{equation}
\begin{aligned}
    v_\pi(s_1) &= -1 + 0.5 v_\pi(s_2) + 0.5 v_\pi(s_3) \\
    v_\pi(s_2) &= 1 + 0.5 v_\pi(s_1) +  0.5 v_\pi(s_4)\\
    v_\pi(s_3) &= 1 + 0.5 v_\pi(s_1) +  0.5 v_\pi(s_4) \\
    v_\pi(s_4) &= 0
\label{23}
\end{aligned}
\end{equation}
Solving the latter, we find 
\begin{equation}
\begin{aligned}
    v_\pi(s_1) &= 0 \quad v_\pi(s_2) &= 1
      \\
    v_\pi(s_3) &= 1  \quad v_\pi(s_4) &= 0
\end{aligned}
\label{23}
\end{equation}

The values of states $ s_2 $ and $ s_3 $ are identical, as both states transition to the terminal state $ s_4 $ in a single step, resulting in the same reward. This equivalence highlights the symmetry in their respective actions and outcomes within the defined environment.

\textbf{Policy Evaluation}: Let us return to the Bellman equation~\eqref{bellman}. We find that the Bellman equations are instrumental in determining the value of states for a specific policy, provided the dynamics of the environment are known. 

Policy evaluation refers to the process of calculating the value function for a given policy using the Bellman equations. Policy evaluation is a crucial step in reinforcement learning that involves calculating the value of each state under a specific policy. By performing policy evaluation, we gain insights into how good the current policy is and whether improvements can be made.

\textbf{Policy Improvement}: Policy improvement refers to the criteria used to enhance a given policy in reinforcement learning. The aim is to find a policy that yields higher expected returns from each state. An improved policy is one where the values of the states, denoted as $ v_\pi(s)$, are greater than those from the previous policy. We will elaborate more on this in Sec.~\ref{Sec8}.

\section{Markov Chain and Markov Decision Process (MDP)}\label{Sec7}

In this section, we examine an important assumption about the environment that leads to transition probabilities to be of the form \( p(s', r \mid s_{1}, a) \). We assume that the probability of transitioning to the next state depends only on the current state and action, not on the agent's previous history. This ``memoryless'' property is called the Markov property, making our environment a Markovian environment.

In other words, given a trajectory of the form 
\begin{align}
    \text{Trajectory} = \left[
(s_{t-2}, a_{t-2}, r_{t-1}),
(s_{t-1}, a_{t-1}, r_{t}),
(s_{t}, a_{t}, r_{t+1}), 
\right. \notag\\
    \left.
(s_{t+1}, a_{t+1}, r_{t+2}),
(s_{t+2}, a_{t+2}, r_{t+3}),
(s_{t+3}, a_{t+3}, r_{t+4})
\right]. \notag
\end{align}
the transition of the agent from state $s_{t}$ to $s_{t+1}$ does not depend on the past history. Mathematically,
\begin{equation}
p(s_{t+1} \mid s_t, s_{t-1}, s_{t-2}) = p(s_{t+1} \mid s_t)
\end{equation}

RL revolves around decision-making. When we operate within an environment that adheres to the Markov property, we are essentially engaging in a Markov Decision Process (MDP). An MDP is defined by a tuple \((\mathcal{S}, \mathcal{A}, P, R, \gamma)\), where:
\begin{itemize}
    \item \(\mathcal{S}\) represents the set of states in the environment,
    \item \(\mathcal{A}\) is the set of possible actions,
    \item \(P\) transition probability matrix,
    \item \(\mathcal{R}\) is the reward function, and
    \item \(\gamma\) is the discount factor, which balances the importance of immediate versus future rewards.
\end{itemize}

\section{Dynamic Programming (DP)}\label{Sec8}

Dynamic programming (DP) is a mathematical optimization method for finding optimal solutions to Markov Decision Processes (MDPs) \cite{Sutton2018-bn}. Using recursive equations, DP efficiently improves policies through techniques like policy iteration and value iteration.

In Sec.~\eqref{objective2}, we discussed about policy evaluation and policy improvement. Let us delve a bit deeper into how policy improvement is implemented and then we will understand policy iteration. Policy improvement aims to find a new, better policy $ \pi' $ than the current policy $ \pi $. The new policy $ \pi' $ is considered better if the value function for all states under $ \pi' $ is at least as high as under $ \pi $. Mathematically, this requires that $ v_{\pi'}(s) \geq v_{\pi}(s) $ for all $ s $ in $ \mathcal{S} $, with strict improvement in at least one state. This means we should seek actions for the agent at each state $ s $  that could improve the value function. The process of seeking the best action for a given state is guided by action-value function $q_{\pi}(s,a)$. The steps to follow are:
\begin{enumerate}
    \item Given the agent is in state $s$, choose a new action $a$ that the agent takes in current state and then follows policy $\pi$. We can, in this case, calculate state-action value $q_{\pi}(s,a)$.
    \item If 
   $$
   q_\pi(s, a) > v_\pi(s),
   $$
   it suggests that taking action $ a $ and then following the current policy is better than just following the policy $ \pi $ all the time from that state.

   \item Update the policy to choose $ a $ in  $s$.

\end{enumerate}

Let us revisit our one-dimensional example illustrated in Figure \ref{fig:rewardfig}, where the policy and transitions are detailed in Sec.~\ref{Sec3}. For state $ s_{6} $, the existing policy prescribes the action \texttt{"right"}, resulting in a state-value of $ -3 $. However, if we were to assign a new action \texttt{"left"} for state $ s_{6} $, we could transition to the terminal state $ s_5 $ (the terminal state) and receive a action-value of $ +3 $. In some case, the new action may not lead to terminal state directly and, in such case, the agent is supposed to follow the old policy until the episode ends. This change indicates that by adopting the new action, we significantly improve the action-value of state $ s_6 $. Consequently, we can update the policy to include this new action for $ s_6 $ as part of our policy improvement process.

The final step is policy iteration. Once we improve a policy $\pi$ to obtain a better policy $\pi'$ using its value function $v_\pi$, we can then find the value of $\pi'$ as $v_{\pi'}$ and improve it further to get a new policy $\pi''$. This creates a chain of increasingly better policies and value functions~\cite{Sutton2018-bn}:
$$
\pi_0 \xrightarrow{\mathrm{E}} v_{\pi_0} \xrightarrow{\mathrm{I}} \pi_1 \xrightarrow{\mathrm{E}} v_{\pi_1} \xrightarrow{\mathrm{I}} \cdots \xrightarrow{\mathrm{I}} \pi_* \xrightarrow{\mathrm{E}} v_*,
$$
where $\xrightarrow{\mathrm{E}}$ represents policy evaluation and $\xrightarrow{\mathrm{I}}$ represents policy improvement. Each step results in a better policy until we reach the optimal policy. 

A drawback of policy iteration is that each iteration requires a complete policy evaluation, which may involve multiple sweeps through the entire state space to achieve convergence. This iterative process can be computationally expensive, especially in environments with a large number of states. In appendices~\ref{AppedixIterMeth} and \ref{AppedixIterPolicy}, we introduce value iteration, an alternative approach that integrates policy evaluation and improvement into a single update step, often resulting in faster convergence.

\section{Monte Carlo Methods}\label{Sec9}

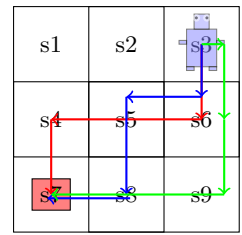
\begin{figure}
    \centering
    \begin{tikzpicture}
        \definecolor{transparentBlue}{rgb}{0.5, 0.5, 1} 
        \definecolor{redCell}{rgb}{1,0.5,0.5}

        \foreach \x in {0,1} {
            \foreach \y in {0,1} {
                \node[draw, minimum size=2cm, align=center] at (\x, -\y) {}; 
            }
        }
        
        \node at (-0.5, 0.5) {s1};
        \node at (0.5, 0.5) {s2};
        \node at (1.5, 0.5) {s3}; 
        \node at (-0.5, -0.5) {s4};
        \node at (0.5, -0.5) {s5}; 
        \node at (1.5, -0.5) {s6};
        \node[fill=redCell, draw] at (-0.5, -1.5) {s7}; 
        \node at (0.5, -1.5) {s8};
        \node at (1.5, -1.5) {s9};

        \draw[fill=transparentBlue, opacity=0.5] (1.3, 0.2) rectangle (1.7, 0.7); 
        \draw[fill=transparentBlue!80, opacity=0.5] (1.4, 0.7) rectangle (1.6, 0.9); 

        \fill[white, opacity=0.5] (1.43, 0.85) circle (0.05); 
        \fill[black, opacity=0.5] (1.43, 0.85) circle (0.02); 
        \fill[white, opacity=0.5] (1.57, 0.85) circle (0.05); 
        \fill[black, opacity=0.5] (1.57, 0.85) circle (0.02); 

        \draw[fill=transparentBlue, opacity=0.5] (1.3, 0.5) rectangle (1.2, 0.55); 
        \draw[fill=transparentBlue, opacity=0.5] (1.7, 0.5) rectangle (1.8, 0.55); 

        \draw[fill=transparentBlue, opacity=0.5] (1.3, 0.2) rectangle (1.4, 0.15); 
        \draw[fill=transparentBlue, opacity=0.5] (1.6, 0.2) rectangle (1.7, 0.15); 

        \draw[red, thick,->] (1.5, 0.5) -- (1.5, -0.5); 
        \draw[red, thick,->] (1.5, -0.5) -- (-0.5, -0.5); 
        \draw[red, thick,->] (-0.5, -0.5) -- (-0.5, -1.5); 

        \draw[blue, thick,->] (1.5, 0.5) -- (1.5, -0.2);
        \draw[blue, thick,->] (1.5, -0.2) -- (0.5, -0.2);
        \draw[blue, thick,->] (0.5, -0.2) -- (0.5, -1.5); 
        \draw[blue, thick,->] (0.55, -1.55) -- (-0.55, -1.55); 

        \draw[green, thick,->] (1.5, 0.5) -- (1.8, 0.5); 
        \draw[green, thick,->] (1.8, 0.5) -- (1.8, -0.5); 
        \draw[green, thick,->] (1.8, -0.5) -- (1.8, -1.5); 
        \draw[green, thick,->] (1.8, -1.5) -- (-0.5, -1.5);
    \end{tikzpicture}
    \caption{Robot's trajectories from cell s3 to the target cell s7 with different paths represented by colored arrows.}
    \label{fig:robot_trajectory}
\end{figure}

Dynamic programming is useful when there is complete knowledge of the environment in the form of transition probabilities. However, most of the time, the environment knowledge is unknown (i.e we known have first hand information about the transition probabilities), and the agent must learn about the environment by interacting with it. In such cases, when knowledge of the environment is missing, we use Monte Carlo methods.

The idea behind Monte Carlo Method in evaluation of the value of states is quite simple.  The idea is that, given any state (assuming the robot or agent is in that state), we generate as many trajectories as possible while following the current policy. For instance, in our \(3 \times 3\) grid, if the agent is at state \(s_3\) and the terminal state is \(s_7\), we have illustrated three possible trajectories (red, blue and green). The rewards is $\mathcal{R} = (-1, 3)$

Once we have these trajectories, the next task is to determine the value of each state. With these trajectories, we can only calculate the values of the states that the trajectories pass through. In our $3 \times 3$ example (see Fig.~\ref{fig:robot_trajectory}), we observe that two trajectories (red and blue) pass through state $s_{5}$ and all the three trajectories pass through state $s_{6}$.

To evaluate the value of state $s_5$, we find the return (sum of future rewards) obtained from each trajectory (red and blue) from $s_5$ and to the the terminal state $s_7$. 
\begin{equation*}
\begin{aligned}
     v_{\pi, red}(s_5) = (-1) + (3) = 2 \\
      v_{\pi, blue}(s_5) = (-1) + (3) = 2 
\end{aligned}
\end{equation*}

Finally, we compute the expected value of $s_5$ by averaging the total rewards across all the trajectories that pass through it:

\begin{equation*}
\begin{aligned}
     v_{\pi, expected}(s_5) = \frac{v_{\pi, red}(s_5) + v_{\pi, blue}(s_5)}{2} = 2
\end{aligned}
\end{equation*}

Similarly, for the state $s_{6}$, we have the following

\begin{equation}
\begin{aligned}
    v_{\pi, \text{red}}(s_6) & = (-1) + (-1) + (3) = 1, \\
    v_{\pi, \text{blue}}(s_6) & = (-1) + (-1) + (3) = 1, \\
    v_{\pi, \text{green}}(s_6) & = (-1) + (-1) + (3) = 1, \\
    v_{\pi, \text{expected}}(s_6) & = \frac{v_{\pi, \text{red}}(s_6) + v_{\pi, \text{blue}}(s_6) + v_{\pi, \text{green}}(s_6)}{3} \\ &= 1
\end{aligned}
\end{equation}

However, it is important to note that we currently have only three trajectories, each corresponding to a unique episode starting from state $s_3$. In practice, it is essential to conduct numerous episodes originating from all states and identify all trajectories that pass through the specific state whose value we wish to calculate. The more episodes we gather, the greater the likelihood that our estimated value will be close to the true value.

The First-visit and Every-visit Monte Carlo methods in MDP are outlined in Appedix~\ref{AppenixFirsEveryZVisit}.

\section{Temporal Difference Methods}\label{Sec10}

Temporal Difference methods was introduced in 1988 by Sutton \cite{Sutton1988, Sutton2018-bn}. In Temporal Difference methods, the learning process takes place through bootstrapping: we update the value of a state based on the estimated value of the next state. We will explain the idea in details later, but before this let us revisit some equations and derive a relation between them. \\
Before we move forward, we introduce the following mathematical relation which will be useful in our future derivation. From Eq \ref{Discount_factor}, we have
\begin{equation}
    G_t \doteq r_{t+1}+\gamma r_{t+2}+\gamma^2 r_{t+3}+\cdots=\sum_{k=0}^{\infty} \gamma^k r_{t+k+1},
\end{equation}
It is easy to see the following relation
\begin{align}
    \begin{split}
        G_t &\doteq r_{t+1} + \gamma r_{t+2} + \gamma^2 r_{t+3} + \gamma^3 r_{t+4} + \cdots \\
            &= r_{t+1} + \gamma G_{t+1}
    \end{split}
    \label{Discount_factor_relation}
\end{align}

Let us revisit Eq.~\eqref{value_function_stochastic} to derive the Bellman equation~\eqref{bellman} from it,

\begin{align}
v_\pi(s) & = \mathbb{E}_\pi\left[G_t \mid S_t = s\right] \notag
\quad\text{(From Eq.~\eqref{value_function_stochastic})}\\
         & = \mathbb{E}_\pi\left[r_{t+1} + \gamma G_{t+1} \mid S_t = s\right] \notag
         \quad\text{(From Eq.~\eqref{Discount_factor_relation})}\\
         & = \sum_a \pi(a \mid s) \sum_{s' \in \mathcal{S} \setminus \{s\}} \sum_{r \in \mathcal{R}} p\left(s^{\prime}, r \mid s, a\right) \notag\\
         & \quad \times \left[r + \gamma \mathbb{E}_\pi\left[G_{t+1} \mid S_{t+1} = s^{\prime}\right]\right] \notag\\
         & = \sum_a \pi(a \mid s) \sum_{\substack{s' \in \mathcal{S} \setminus \{s_3\} \\ r \in \mathcal{R}}} p\left(s^{\prime}, r \mid s, a\right) \notag\\
         & \quad \times \left[r + \gamma v_\pi\left(s^{\prime}\right)\right], \quad \text{for all } s \in \mathcal{S}.
    \label{bellman_derivation}
\end{align}

where we have used $v_\pi(s')  = \mathbb{E}_\pi\left[G_{t+1} \mid S_{t+1} = s'\right]$

The Bellman equation states that, given a policy $ \pi(a \mid s) $ and the environment dynamics $ p(s^{\prime}, r \mid s, a) $, we can determine the value of any state $ s \in \mathcal{S} $ if the values of all possible subsequent states $ s' \in \mathcal{S} \setminus \{s\} $ are known. In other words, if we know the values $ v(s') $ for all reachable states $ s' $, we can compute the value $ v(s) $ by solving the right-hand side of the Bellman equation. For example, in Eq.~\eqref{23}, if $ v_{\pi} (s_2) $ is unknown but the values of other states are known, then using these known state values, we can find $ v_{\pi} (s_2) $ through the Bellman equation. In Eq.~\eqref{bellman_derivation}, an intermediate step is as follows:
\begin{equation}
    \begin{aligned}
         v_\pi(s) & = \mathbb{E}_\pi\left[G_t \mid S_t = s\right] \\
         & =  \mathbb{E}_\pi\left[r_{t+1} + \gamma v_{\pi} (S_{t+1}) \mid S_t = s\right]
    \end{aligned}
    \label{state-equation-state}
\end{equation}

We must note that the action $ a $ is determined by the policy structure $ \pi $, which defines the agent’s behavior in choosing actions. However, the reward $ r $ received and the next state $ s' $ reached after taking action  $a$ are governed by the environment's dynamics. These dynamics are encapsulated in the environment’s transition function \( p(s', r \mid s, a) \), which specifies the probability distribution over possible next states and rewards given the current state-action pair.

To interpret Eq.~\eqref{state-equation-state}, let us consider a deterministic environment.This implies that the agent's trajectory is unique; regardless of the number of episodes, the trajectory remains consistent, eliminating any stochasticity in state transitions. Consequently, we can omit the expectation term from the equation. Under these conditions, the equation suggests that, given the agent follows the policy $ \pi $, the value of the current state $ s $ is equal to the sum of the reward $ r_{t+1} $ obtained after taking action $ a$ and the discounted value of the subsequent state $ S_{t+1} $ where the agent transitions. 
\begin{equation}
    \begin{aligned}
         v_\pi(s) = r_{t+1} + \gamma v_{\pi} (S_{t+1}), \quad \text{given} \quad S_t = s
    \end{aligned}
    \label{state-equation-state-1}
\end{equation}
This provides a recursive relationship in which the value of each state can be computed based on the immediate reward and the value of the succeeding state. 

Let us change track and introduce an interesting update rule for the state value as follows:
\begin{equation}
    \begin{aligned}
         v_\pi(s) & \leftarrow v_\pi(s) + \alpha \left( r_{t+1} + \gamma v_{\pi}(S_{t+1}) - v_\pi(s) \right), \\
         &\quad \text{given} \quad S_t = s, \, S_{t+1} = s'
    \end{aligned}
    \label{state-equation-state-2}
\end{equation}
where $\alpha$ is the learning rate. \\
The update rule in Eq.~\eqref{state-equation-state-2} suggests that the value of a given state $ s $ can be adjusted immediately by considering the expected return from the next state $ S_{t+1} $ in terms of the immediate reward $r_{t+1}$ and the value $ v_{\pi}(S_{t+1}) $. This incremental approach allows for more efficient updates, as it does not require waiting until the end of an episode to evaluate the entire trajectory, in contrast to Monte Carlo methods where updates are only applied at the episode's conclusion. Instead, this rule leverages a technique called temporal difference learning. Temporal difference methods update the value of a state by comparing the current estimate \( v_\pi(s) \) with a new estimate based on the observed outcome, specifically \( r_{t+1} + \gamma v_\pi(S_{t+1}) \). This difference $\left( r_{t+1} + \gamma v_{\pi}(S_{t+1}) - v_\pi(s) \right)$ is known as the temporal difference error.

Let us illustrate this using the 2D grid example in Fig.~\ref{fig:robot_trajectory}. Let
the initial value of all the states be zero, $v_{\pi}(s_1) = v_{\pi}(s_2) = \ldots = v_{\pi}(s_9)= 0$. Given policy $\pi$, the agent in state $s = s_3$ moves down to state $s' = s_6$, following the green trajectory. In this process, it receives a reward of $-1$. Thus, the updated value of state $s_3$, following Eq.~\eqref{state-equation-state-2}, is given by ($\alpha = 0.1, \gamma = 0.9$)
\begin{equation}
\begin{aligned}
    v_\pi(s_3) &\leftarrow v_\pi(s_3) + \alpha \left( r_{t+1} + \gamma v_\pi(s_6) - v_\pi(s_3) \right) \\
    &\implies v_\pi(s_3) \leftarrow 0 + 0.1 \left( -1 + 0.9 \cdot 0 - 0 \right) \\
    &\implies v_\pi(s_3) \leftarrow 0 + 0.1 \cdot (-1) \\
    &\implies v_\pi(s_3) \leftarrow -0.1 \\
    &\implies v_\pi(s_3) = -0.1
\end{aligned}
\end{equation}
Let us recall that the primary objective in Temporal Difference learning is to estimate the correct value of each state. Unlike Monte Carlo methods, where the value of a state is updated only at the end of an episode, after observing the full trajectory and accumulating all rewards, Temporal Difference methods update the state value immediately following each action. In here, the update of a state’s value happens each time the state  is visited, even if revisited within a single trajectory. 

For example, in Monte Carlo, the value of a state, say $ s_3 $, is adjusted only after the episode concludes (follow the green trajectory) and the agent's path has yielded all relevant rewards. In Temporal Difference, however, the update occurs right after each transition, based on the immediate reward and the estimated value of the subsequent state. This approach enables incremental updates across multiple trajectories, leveraging each within-episode transition for faster convergence and improved sample efficiency.

Further, building on the concept of one-step Temporal Difference learning, we can explore two key Temporal Difference based algorithms: SARSA State – Action – Reward – State – Action) and Q-learning. The fundamental difference lies in the update method.  SARSA, an on-policy method, updates the value of a state-action pair by considering the actual next action taken by the agent, using the following rule on action value function:
\begin{equation}
    q(s, a) \leftarrow q(s, a) + \alpha (r_{t+1} + \gamma q(s', a') - q(s, a))
\end{equation}
 Q-learning, an off-policy method, instead updates the value by looking at the highest possible action in the next state, with the rule:
\begin{equation}
    q(s, a) \leftarrow q(s, a) + \alpha (r_{t+1} + \gamma \max_{a'} q(s', a') - q(s, a))
\end{equation}

This difference allows SARSA to learn values based on the agent's actual actions, while Q-learning aims for the best possible actions regardless of the policy followed.

\section{Policy Based Methods}\label{Sec11}

In this section, we will try something new, but our goal stays the same, finding the best possible policy. Let us quickly recall what we did before. Previously, to get an optimal policy, we first found the value of each action. Then, using these action values, we decided which action was best in every state and put together our final policy. But now let us think differently. Can we skip the step of calculating action-values altogether? Surprisingly, the answer is yes.

In this new approach, we follow a different path to discover the optimal policy. Instead of using a fixed policy $ \pi(a \mid s) $, where the agent selects actions based on predefined rules or values, we now define the policy as a parameterized function $ \pi(a \mid s; \theta) $. The parameter $ \theta $ governs the behavior of the policy and can be adjusted through learning. This makes the policy not only flexible but also trainable. By tuning $ \theta $, the agent can gradually improve its performance according to experience. Further, we can extend this formulation to incorporate multiple trainable parameters by introducing a parameter vector $\boldsymbol{\theta} = [\theta_1, \theta_2, \ldots, \theta_n]$, which allows for a more expressive representation of the policy in gradient-based reinforcement learning approaches. In such case, the policy can be represented as $\pi(a \mid s;\boldsymbol{\theta})$.

This method stands in contrast to the value-based approach, where the policy is improved by first estimating action-values and then selecting actions that maximize those values. In the policy gradient framework, the policy is improved directly through changes in the parameters $\boldsymbol{\theta}$, without relying on explicit value estimates. This allows for a more direct and often more scalable way to optimize the agent's behavior, especially in complex or continuous action spaces.

\subsection{Policy Gradient I: The Theory} \label{Policy_gradient}

In policy gradient methods, our goal is to directly optimize the performance of a parameterized policy,
$\pi(a\mid s;\boldsymbol{\theta})$. Let us define a scalar objective function $ J(\boldsymbol{\theta} ) $ that measures the quality of the policy. Specifically, $ J(\boldsymbol{\theta})$ is defined as the \textit{expected return} when the agent starts interacting with the environment at the initial step $ t = 0 $ and follows the policy $ \pi(a \mid s;\boldsymbol{\theta}) $ (in short notation $\pi_{\boldsymbol{\theta}}$) thereafter,
\begin{equation}
   J(\boldsymbol{\theta} ) \doteq \mathbb{E}_{\pi_{\boldsymbol{\theta}}}[G_t],  \quad \text{with } t = 0. 
   \label{J}
\end{equation}
remember that we expected return because it provides a stable and reliable measure of a policy's average performance in a stochastic environment (refer back to State Value Function subsection in Sec.~\ref{Sec3}).

Let us understand how Eq.~\eqref{J} is different from our earlier definition $v_{\pi}(s) = \mathbb{E}_{\pi} [G_t | S_t = s]$ in Eq.~\eqref{value_function_stochastic}. The key distinction lies in the conditioning of the expectation. The expression $ v_{\pi}(s) $ represents the expected return starting from a specific state $ s $, and thus captures how desirable it is to be in that particular state under policy $\pi$. In contrast, $ J(\boldsymbol{\theta}) $ evaluates the expected return starting from the initial time step $ t = 0 $. Therefore, $ J(\boldsymbol{\theta}) $ provides a global measure of the overall performance of the policy, averaged across entire episodes.

Now that policy $\pi_{\boldsymbol{\theta}}$ is defined as a differentiable function of parameters $\boldsymbol{\theta}$ and the objective $J(\boldsymbol{\theta})$, policy gradient methods aim to improve policy by maximizing $J(\boldsymbol{\theta})$. This is done by computing the gradient $\nabla_{\boldsymbol{\theta}} J(\boldsymbol{\theta})$, which indicates the direction in which the parameters should be adjusted to increase the expected return. Say, $\boldsymbol{\theta}$ is a vector of $d$ parameters, then the gradient $\nabla_{\boldsymbol{\theta}} J(\boldsymbol{\theta})$ is itself a vector, given by
\begin{equation}
\nabla_{\boldsymbol{\theta}} J(\boldsymbol{\theta}) = \begin{bmatrix} 
\frac{\partial J}{\partial \theta_1} \\ 
\frac{\partial J}{\partial \theta_2} \\ 
\vdots \\ 
\frac{\partial J}{\partial \theta_d} 
\end{bmatrix}.
\end{equation}

The parameters are then updated using gradient ascent:
\begin{equation}
\boldsymbol{\theta}_{k+1} = \boldsymbol{\theta}_k + \alpha \nabla_{\boldsymbol{\theta}} J(\boldsymbol{\theta})\Big|_{\boldsymbol{\theta}=\boldsymbol{\theta}_k},
\label{updaterule}
\end{equation}
where \( \alpha \) is the learning rate. In this way, the policy is directly optimized without requiring a value function estimate.

Note that, the gradient $\nabla_{\boldsymbol{\theta}} J(\boldsymbol{\theta})$ is a vector of partial derivatives, one for each parameter $\theta_i$. Thus, the update rule in Eq.~\eqref{updaterule} adjusts all parameters simultaneously:
\begin{equation}
\theta_{i,k+1} = \theta_{i,k} + \alpha \, \frac{\partial J}{\partial \theta_i}\Big|_{\boldsymbol{\theta}=\boldsymbol{\theta}_k}, \quad \text{for } i = 1, \ldots, d.
\end{equation}
In other words, every parameter is pulled in the direction that increases the objective, and the step size is controlled by the learning rate $\alpha$.

\subsection{Policy Gradient II: Example}

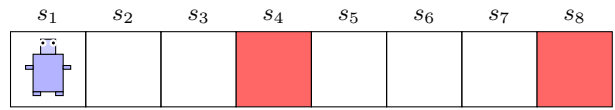
\begin{figure}
    \centering
    \begin{tikzpicture}
        \foreach \x in {1, 2, ..., 8} {
            \ifnum\x=4
                \draw[fill=red!60] (\x, 0) rectangle (\x + 1, 1); 
            \else
                \ifnum\x=8
                    \draw[fill=red!60] (\x, 0) rectangle (\x + 1, 1); 
                \else
                    \draw (\x, 0) rectangle (\x + 1, 1); 
                \fi
            \fi
            \node at (\x + 0.5, 1.2) {$s_\x$};
        }

        \draw[fill=blue!30] (1.3, 0.2) rectangle (1.7, 0.7); 
        \draw[fill=blue!20] (1.4, 0.7) rectangle (1.6, 0.9); 
        \fill[white] (1.43, 0.85) circle (0.05); 
        \fill[black] (1.43, 0.85) circle (0.02);
        \fill[white] (1.57, 0.85) circle (0.05); 
        \fill[black] (1.57, 0.85) circle (0.02);
        \draw[fill=blue!30] (1.3, 0.5) rectangle (1.2, 0.55); 
        \draw[fill=blue!30] (1.7, 0.5) rectangle (1.8, 0.55); 
        \draw[fill=blue!30] (1.3, 0.2) rectangle (1.4, 0.15); 
        \draw[fill=blue!30] (1.6, 0.2) rectangle (1.7, 0.15); 
    \end{tikzpicture}
    \caption{Grid world with robot in $s_1$ and terminal states $s_4$ and $s_8$ highlighted in red}
    \label{fig:policy_gradient_0}
\end{figure}

Let us understand how we can implement the formulation of the policy gradient for the 1D grid example (Fig.~\ref{fig:policy_gradient_0}). The example in Fig.~\ref{fig:grid_with_robot_and_wall} was easy for the agent to learn. The robot just needed to go left if the terminal state was on its left, and right if it was on its right. The new example in Fig.~\ref{fig:policy_gradient_0} is more challenging because it has two terminal states $s_4$ and $s_8$. In state $s_5$, the robot should always go left. In state $s_6$, both left and right lead to terminal states, so the agent needs to choose wisely. In state $s_7$, the best action is again to go right. This setup shows that learning becomes harder when the environment has more complex decisions.

If we were to implement Monte Carlo or Temporal difference method, we could start with the following random policy:

\begin{lstlisting}
policy = {
    1: {'right': 1.0},                
    2: {'left': 0.5, 'right': 0.5},   
    3: {'left': 0.5, 'right': 0.5},   
    4: {},                          
    5: {'left': 0.5, 'right': 0.5},   
    6: {'left': 0.5, 'right': 0.5},   
    7: {'left': 0.5, 'right': 0.5},  
    8: {}                            
}
\end{lstlisting}

The initial random policy must be improved to reach a stable optimal policy. However, since we are working within the policy gradient framework, it is essential to define a trainable parameter, denoted by $\theta$, or more generally as a parameter vector $\boldsymbol{\theta} = [\theta_1, \theta_2, \ldots, \theta_6]$. Consequently, we must revise our initial policy representation from $\pi(a \mid s)$ to the parameterized form $\pi(a \mid s; \boldsymbol{\theta})$ to reflect this dependency. A simple way to do this is as follows (one must remember that the dependency on $\theta$ can be chosen in a more complex form):
\[
\pi(a \mid s;\theta) \;=\;
\begin{array}{|c|c|c|}
\hline
s & \pi(left\mid s;\theta) & \pi(right\mid s;\theta) \\
\hline
1 & 0 & 1-\theta \\
2 & 0.5+\theta & 0.5-\theta \\
3 & 0.5+\theta & 0.5-\theta \\
4 & 0 & 0 \\
5 & 0.5+\theta & 0.5-\theta \\
6 & 0.5+\theta & 0.5-\theta \\
7 & 0.5+\theta & 0.5-\theta \\
8 & 0 & 0 \\
\hline
\end{array}
\]

or in more general form


\[
\pi(a \mid s;\theta) \;=\;
\begin{array}{|c|c|c|}
\hline
s & \pi(left\mid s;\theta) & \pi(right\mid s;\theta) \\
\hline
1 & 0 & 1-\theta_1 \\
2 & 0.5+\theta_2 & 0.5-\theta_2 \\
3 & 0.5+\theta_3 & 0.5-\theta_3 \\
4 & 0 & 0 \\
5 & 0.5+\theta_4 & 0.5-\theta_4 \\
6 & 0.5+\theta_5 & 0.5-\theta_5 \\
7 & 0.5+\theta_6 & 0.5-\theta_6 \\
8 & 0 & 0 \\
\hline
\end{array}
\]

The general policy gradient update is given by Eq.~\eqref{updaterule},
where the goal is to adjust the policy parameters $\boldsymbol{\theta}$ in the direction that maximizes the expected return $J(\boldsymbol{\theta})$. However, directly computing $\nabla_{\boldsymbol{\theta}} J(\boldsymbol{\theta})$ is difficult because it involves expectations over all possible trajectories. To address this, the REINFORCE algorithm\cite{williams1992simple} uses a clever trick: it approximates this gradient using sampled trajectories and the log-likelihood trick. Specifically, it substitutes the gradient with
\begin{equation}
\nabla_{\boldsymbol{\theta}} J(\boldsymbol{\theta}) \approx G_t \cdot \nabla_{\boldsymbol{\theta}} \log \pi(a_t \mid s_t; \boldsymbol{\theta}),
\label{reinf}
\end{equation}
where $G_t$ is the return from the time step $t$ onward. This leads to the \textit{reinforce} update rule:

\begin{equation}
\boldsymbol{\theta} \leftarrow \boldsymbol{\theta} + \alpha \cdot G_t \cdot \nabla_{\boldsymbol{\theta}} \log \pi(a_t \mid s_t; \boldsymbol{\theta}).
\end{equation}

\begin{figure}
    \centering
    \begin{tikzpicture}
        \foreach \x in {1, 2, ..., 8} {
            \ifnum\x=4
                \draw[fill=red!60] (\x, 0) rectangle (\x + 1, 1); 
            \else
                \ifnum\x=8
                    \draw[fill=red!60] (\x, 0) rectangle (\x + 1, 1); 
                \else
                    \draw (\x, 0) rectangle (\x + 1, 1); 
                \fi
            \fi
            \node at (\x + 0.5, 1.2) {$s_\x$};
        }

        \draw[fill=blue!30] (1.3, 0.2) rectangle (1.7, 0.7); 
        \draw[fill=blue!20] (1.4, 0.7) rectangle (1.6, 0.9); 
        \fill[white] (1.43, 0.85) circle (0.05); \fill[black] (1.43, 0.85) circle (0.02);
        \fill[white] (1.57, 0.85) circle (0.05); \fill[black] (1.57, 0.85) circle (0.02);
        \draw[fill=blue!30] (1.3, 0.5) rectangle (1.2, 0.55);
        \draw[fill=blue!30] (1.7, 0.5) rectangle (1.8, 0.55);
        \draw[fill=blue!30] (1.3, 0.2) rectangle (1.4, 0.15);
        \draw[fill=blue!30] (1.6, 0.2) rectangle (1.7, 0.15);

        \draw[->, thick] (2.5, 0.5) -- (3.5, 0.5) node[midway, above] {\footnotesize $0.99$}; 
        \draw[->, thick] (3.5, 0.5) -- (4.5, 0.5) node[midway, above] {\footnotesize $0.99$}; 
        \draw[->, thick] (1.5, 0.5) -- (2.3, 0.5) node[midway, above] {\footnotesize $1$}; 

        \draw[->, thick] (5.5, 0.5) -- (4.5, 0.5) node[midway, above] {\footnotesize $0.99$}; 
        \draw[->, thick] (6.5, 0.5) -- (7.5, 0.5) node[midway, above] {\footnotesize $0.99$}; 
        \draw[->, thick] (7.5, 0.5) -- (8.5, 0.5) node[midway, above] {\footnotesize $0.99$}; 
    \end{tikzpicture}
    \caption{Improved policy with action directions and probabilities derived from the policy gradient method. Terminal states $s_4$ and $s_8$ are highlighted in red. The numerical values indicate the transition probabilities for each action, demonstrating the learned stochastic policy after convergence.}
    \label{fig:policy_gradient}
\end{figure}
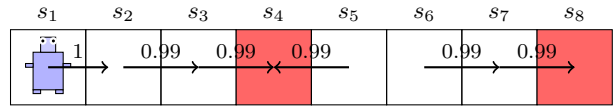

This formulation allows learning from actual sampled episodes by increasing the probability of actions that lead to higher returns (see Fig.~\ref{fig:policy_gradient}), even without knowing the full model of the environment.

Let us now look at the improved policy  for Fig.~\ref{fig:policy_gradient_0},
\begin{lstlisting}
    improved_policy = {
    "s1": {'right': 1.00, 'left': 0.00},                
    "s2": {'right': 0.99, 'left': 0.01},   
    "s3": {'right': 0.99, 'left': 0.01},   
    "s4": {},  # Terminal state
    "s5": {'right': 0.01, 'left': 0.99},  
    "s6": {'right': 0.99, 'left': 0.01},   
    "s7": {'right': 0.99, 'left': 0.01},  
    "s8": {}  # Terminal state
}

\end{lstlisting}

\section{Actor-Critic Method}
\label{xii}
Policy gradient methods, as introduced earlier in Sec. XI, work quite well in environments where rewards are sparse (agent only receive reward at the end of an episode), such as the 1D grid world. In those settings, the agent learns by seeing the total result of an episode, which helps guide its future decisions.

However, for long episodes in sparse reward settings, one of the critical issues is that they suffer from high variance in their gradient estimates. A simple policy gradient algorithm updates its policy based on the total return of an entire episode. The problem is that the episode's total reward can vary wildly due to randomness, even if the agent's actions are similar. This makes the gradient updates noisy and unstable, leading to slow and inefficient learning.\cite{greensmith2004variance, rl_flaws_tds, peters2008natural}.

Also, policy gradient is a very slow learning method like Monte Carlo.\cite{wang2020long_horizon_rl,zhang2021sample_efficient_reinforce}. This is because they typically wait until the entire episode is complete before making any updates.And that is why, in environments with long tasks or many steps (known as long-horizon environments), this leads to inefficient and unstable training.

To solve these problems, the \textit{actor-critic method}\cite{barto1983actorcritic} offers a useful solution. It combines the strengths of both value-based methods (like Temporal Difference) and policy-based methods (like policy gradient). As the name suggests:
\begin{itemize}
    \item The \textit{actor} is the part that \textit{decides what to do} it updates the policy $\pi(a|s; \theta)$, just like in regular policy gradient.
    \item The \textit{critic} is the part that \textit{evaluates how good a situation is}—it uses a value function $V(s; w)$, learned using temporal difference techniques.
\end{itemize}
  
The policy gradient update rule (introduced in Sec.~XI) is modified as follows:

\begin{equation}
\nabla_{\theta} J(\theta) \approx A(s_t, a_t) \cdot \nabla_{\theta} \log \pi_{\theta}(a_t \mid s_t) 
\label{eq:policy_gradient}
\end{equation}

where:
\begin{itemize}
    \item $J(\theta)$ is the performance objective, i.e., the expected total reward.
    \item $\pi_{\theta}(a_t \mid s_t)$ is the probability of taking action $a_t$ in state $s_t$ under the policy parameterized by $\theta$.
    \item $\nabla_{\theta} \log \pi_{\theta}(a_t \mid s_t)$ represents how a small change in $\theta$ affects the log-probability of the chosen action.
    \item $A(s_t, a_t)$ is the \emph{advantage function}, which measures how much better or worse action $a_t$ was compared to the critic’s baseline expectation for state $s_t$.
\end{itemize}

In the actor--critic algorithm, the advantage function \(A(s_t, a_t)\) serves as a replacement for the return \(G_t\) used in Equation.~\eqref{reinf} of the REINFORCE policy gradient method. The advantage function is often estimated using the \emph{temporal difference error}:

\begin{equation}
A(s_t, a_t) \approx \delta_t = r_{t+1} + \gamma V(s_{t+1}; w) - V(s_t; w),
\label{eq:td_error}
\end{equation}

where:
\begin{itemize}
    \item $r_{t+1}$ is the immediate reward received after taking action $a_t$ in state $s_t$,
    \item $\gamma \in [0,1]$ is the discount factor determining the importance of future rewards,
    \item $V(s_t; w)$ is the critic’s estimate of the value of state $s_t$,
    \item $V(s_{t+1}; w)$ is the critic’s estimate of the value of the next state.
\end{itemize}

The temporal difference error shows how surprising the outcome is. 
If the reward we receive, along with the value of the next state, 
is greater than expected, then $\delta_t > 0$. 
If it is smaller than expected, then $\delta_t < 0$.

So now there will be two updates. 
The critic updates its value function parameters $w$ according to:

\begin{equation}
w \leftarrow w + \beta \, \delta_t \, \nabla_{w} V(s_t; w),
\label{eq:critic_update}
\end{equation}

where $\beta > 0$ is the critic’s learning rate, and $\nabla_{w} V(s_t; w)$ is the gradient of the value function with respect to its parameters.  
This update moves the critic’s predictions closer to the observed outcomes.

Finally, the actor updates its policy parameters $\theta$ using the same temporal difference error:

\begin{equation}
\theta \leftarrow \theta + \alpha \, \delta_t \, \nabla_{\theta} \log \pi_{\theta}(a_t \mid s_t),
\label{eq:actor_update}
\end{equation}

where $\alpha > 0$ is the actor’s learning rate.  
\subsection*{Actor-Critic Update Example on 1D Grid}

Let us take the 1D grid example. Initially we have,
\[
\begin{array}{rl}
\texttt{$\pi(a \mid s;\boldsymbol{\theta})$ = \{} \\
\texttt{1:} & \texttt{\{right: } 1 - \theta_1 \texttt{\},} \\
\texttt{2:} & \texttt{\{left: } 0.5 + \theta_2 \texttt{, right: } 0.5 - \theta_2 \texttt{\},} \\
\texttt{3:} & \texttt{\{left: } 0.5 + \theta_3 \texttt{, right: } 0.5 - \theta_3 \texttt{\},} \\
\texttt{4:} & \texttt{\{\},} \\
\texttt{5:} & \texttt{\{left: } 0.5 + \theta_4 \texttt{, right: } 0.5 - \theta_4 \texttt{\},} \\
\texttt{6:} & \texttt{\{left: } 0.5 + \theta_5 \texttt{, right: } 0.5 - \theta_5 \texttt{\},} \\
\texttt{7:} & \texttt{\{left: } 0.5 + \theta_6 \texttt{, right: } 0.5 - \theta_6 \texttt{\},} \\
\texttt{8:} & \texttt{\{\}} \\
\texttt{\}} &
\end{array}
\]
All policy parameters are initialized to zero: $\theta_1 = \theta_2 = \dots = \theta_6 = 0 \quad$

Here learning rate $\alpha = 0.1$ (for both actor and critic update) and discount factor $\gamma = 0.9$. The initial value function is $v(s)=0$ for all states.

Consider the trajectory$s_6 \rightarrow s_7 \rightarrow s_8 \ (\text{terminal})$. The temporal difference error and critic update are:
\begin{equation}
\delta_t = r_t + \gamma v(s_{t+1}) - v(s_t), \quad
v(s_t) \leftarrow v(s_t) + \alpha \delta_t
\end{equation}

Transition $s_6 \to s_7$ (reward $-1$):
\begin{align*}
v(s_6) &= 0, \quad v(s_7)=0 \\
\delta &= -1 + 0.9\cdot 0 - 0 = -1 \\
v(s_6) &\leftarrow 0 + 0.1 \cdot (-1) = -0.1
\end{align*}

Transition $s_7 \to s_8$ (reward $+5$, terminal $s_8$):
\begin{align*}
v(s_7) &= 0, \quad v(s_8)=0 \\
\delta &= 5 + 0.9\cdot 0 - 0 = 5 \\
v(s_7) &\leftarrow 0 + 0.1 \cdot 5 = 0.5
\end{align*}

After critic updates:
\[
v(s_6)=-0.1, \quad v(s_7)=0.5, \quad v(s_8)=0
\]

Next step is to calculate the advantage function,
\begin{equation}
A(s_t,a_t) = r_t + \gamma v(s_{t+1}) - v(s_t)
\end{equation}

- For $s_6 \to s_7$:
\[
A(s_6,right) = -1 + 0.9 \cdot 0.5 - (-0.1) = -0.45
\]

- For $s_7 \to s_8$:
\[
A(s_7,right) = 5 + 0.9 \cdot 0 - 0.5 = 4.5
\]

The actor update is:
\begin{equation}
\theta \leftarrow \theta + \alpha \, A(s_t,a_t)\, \nabla_\theta \log \pi_\theta(a_t \mid s_t)
\end{equation}

With the linear parametrization:
\[
\pi(right \mid s) = 0.5 - \theta, \qquad 
\pi(left \mid s) = 0.5 + \theta
\]
the derivative is:
\[
\frac{\partial}{\partial \theta} \log \pi(right \mid s) = -\frac{1}{0.5 - \theta}
\]

Update at $s_6$ (action = right, $\theta_6=0$):
\begin{align*}
\Delta \theta_6 &= \alpha \, A(s_6,right) \left(-\frac{1}{0.5-\theta_6}\right) \\
&= 0.1 \times (-0.45) \times (-2) \\
&= 0.09 \\
\theta_6' &= 0 + 0.09 = 0.09 \\
\pi'(right \mid s_6) &= 0.5 - 0.09 = 0.41 \\
\pi'(left \mid s_6) &= 0.5 + 0.09 = 0.59
\end{align*}

Update at $s_7$ (action = right, $\theta_7=0$):
\begin{align*}
\Delta \theta_7 &= 0.1 \times 4.5 \times (-2) = -0.9 \\
\theta_7' &= 0 - 0.9 = -0.5 \quad (\text{clipped to } [-0.5,0.5]) \\
\pi'(right \mid s_7) &= 0.5 - (-0.5) = 1.0 \\
\pi'(left \mid s_7) &= 0.0
\end{align*}

 Final Values and Policy:

\begin{align*}
v(s_6) &= -0.1, \quad v(s_7) = 0.5, \quad v(s_8) = 0 \\
\pi(right \mid s_6) &= 0.41, \quad \pi(left \mid s_6)=0.59 \\
\pi(right \mid s_7) &= 1.0, \quad \pi(left \mid s_7)=0.0
\end{align*}

These updates illustrate how the actor-critic framework adjusts the value estimates and policy probabilities step by step along a trajectory in a 1D grid world.

\section{Quantum Control: Introduction}
\label{xiv}
The concept of quantum control emerged in the mid-1980s, when it was realized that it might be possible to deliberately guide the behavior of quantum systems using carefully designed electromagnetic fields. In quantum mechanics, atoms and molecules can reach the same final state through multiple quantum pathways, and these pathways can interfere with each other much like overlapping ripples on water. It was demonstrated~\cite{tannor1985, brumer1986, judson1992teaching} that by shaping the phase and timing of laser pulses, one could exploit this interference to control chemical reactions and molecular transitions. This insight laid the foundation for what became known as \textit{coherent quantum control}. Subsequent theoretical and experimental studies expanded this idea into a general framework for manipulating atomic and molecular dynamics~\cite{tannor1985, warren1993}.

At first, the theoretical framework remained largely conceptual because probing and manipulating quantum systems at such fine scales was extremely difficult. These ideas could not be realized experimentally until the 1990s, which marked a period of rapid experimental progress enabled by the development of femtosecond laser technology and pulse-shaping techniques. These advances made it possible to synthesize and control ultrashort laser pulses on timescales comparable to molecular vibrations and electronic transitions, transforming coherent control from a theoretical concept into an experimentally realizable discipline~\cite{assion1998, herek2002}. 

At the same time, significant progress was made on the theoretical and computational side of quantum control. Researchers developed \textit{Quantum Optimal Control Theory}, a framework that combines the time-dependent Schrödinger equation with optimization methods to design external control fields that achieve specific goals, such as transferring population between states, implementing high-fidelity quantum gates, or reducing decoherence~\cite{peirce1988}. A major milestone was the introduction of the GRAPE algorithm (Gradient Ascent Pulse Engineering) by Khaneja et~al.~(2005), which provided a fast and efficient way to solve high-dimensional optimization problems. Originally developed for nuclear magnetic resonance (NMR) systems, GRAPE was later adapted for use in quantum information processing~\cite{khaneja2005}. Since then, improved variants such as Krotov’s method~\cite{krotov1996}, CRAB (Chopped Random Basis)~\cite{caneva2011}, and GOAT (Gradient Optimization of Analytic Controls)~\cite{egger2014} have further extended the computational tools available for designing precise quantum control fields~\cite{machnes2018}. Recently, globally optimal quantum control methods were developed~\cite{bondar_globally_2025, gaggioli2025unitary}.

By the 2010s, research in quantum control increasingly focused on quantum technologies, including superconducting qubits, trapped ions, and solid-state spin systems. In these platforms, highly accurate and noise-resilient control is essential for building high-fidelity quantum gates and achieving scalable quantum computation~\cite{glaser2015}. The growing complexity of large, open quantum systems exposed to environmental noise motivated researchers to explore machine learning (ML) approaches to automate and improve control design. ML-based quantum control leverages data-driven models, neural networks, and gradient-based optimization to learn control strategies that are adaptive, robust, and efficient for practical hardware~\cite{genois}.


The field of Quantum Reinforcement Learning(QRL) combines ideas from control theory, quantum dynamics, and artificial intelligence. In QRL, controlling a quantum system is formulated as a sequential decision-making process. We have already learned the components of RL, in this case the quantum system acts as the environment, the applied control pulses represent the actions, and performance metrics such as fidelity or energy cost serve as the reward function. Reinforcement learning (RL) agents can therefore learn optimal control strategies either by interacting directly with experiments (model-free) or by training on simulations of quantum dynamics (model-based). 

Recent studies have demonstrated that deep RL can effectively perform tasks such as quantum state preparation, gate optimization, and error suppression even in the presence of environmental noise and decoherence~\cite{bukov2018,zheng2019,GIANNELLI2022,gao2022,koutromanos2024,ernst2025,robustQRL2025}. With continued progress in quantum hardware and hybrid quantum-classical computing, QRL is a promising direction toward autonomous, adaptive, and robust controllers for next-generation quantum technologies.


\section{Application of RL in Quantum Control}
\label{apprl}

The goal of optimal terminal control is to drive the system from a known initial quantum state to a desired target state at a specified terminal (i.e., final) time \( T \). The evolution of the quantum system is governed by the time-dependent Schrödinger equation:
\begin{equation}
    i\hbar \frac{d\ket{\psi(t)}}{dt} = H(u(t)) \ket{\psi(t)},
\end{equation}
where \( \ket{\psi(t)} \) is the quantum state at time \( t \), and \( H(u(t)) \) is the Hamiltonian, which depends on the control field \( u(t) \).

To quantify the success of the control, we use the \emph{fidelity} between the final state \( \ket{\psi(T)} \) and the desired target state \( \ket{\psi_{\text{target}}} \):
\begin{equation}
F = \left| \langle \psi_{\text{target}} | \psi(T) \rangle \right|^2.
\end{equation}
A fidelity \( F = 1 \) indicates perfect overlap with the target state.

\subsection{An Example of Simple Optimal control Problem}



\begin{figure*}[t]
    \centering
    \begin{subfigure}{0.45\textwidth}
        \includegraphics[width=\textwidth]{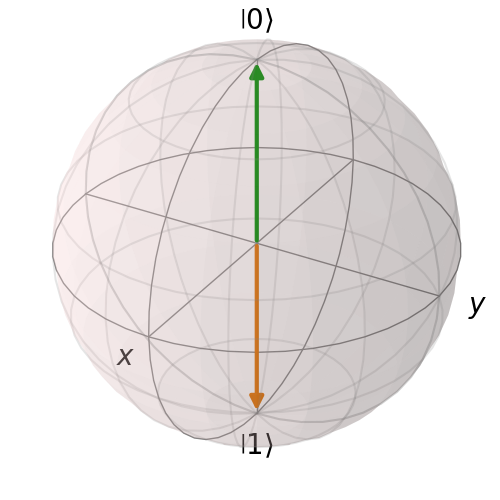}
        \caption*{(a)}
    \end{subfigure}
    \begin{subfigure}{0.45\textwidth}
        \includegraphics[width=\textwidth]{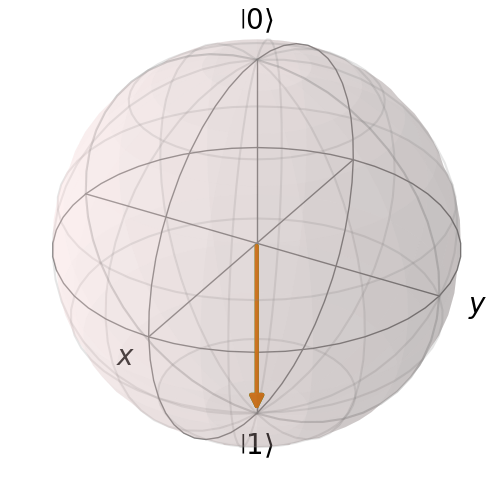}
        \caption*{(b)}
    \end{subfigure}
    
    \caption{
        Bloch sphere representations of quantum state evolution.  
        (\textbf{a}) The initial state (in green) corresponds to the computational basis state $\ket{0}$, and the target state (in orange) is $\ket{1}$.  
        (\textbf{b}) The final state after control (in green) overlaps with the target state (in orange), indicating successful evolution. The overlap causes only one visible vector due to their alignment.
    }
    \label{blooch}
\end{figure*}


Consider a single qubit, initially in state \( |0\rangle \), and we aim to drive it to the target state \( |1\rangle \) using a control field. The control is applied through a rotation operator along the X-axis, which is parameterized by a rotation angle \( \theta \).

\begin{itemize}
    \item {Initial state:} \( |0\rangle \)
    \item {Target state:} \( |1\rangle \)
    \item {Control parameter:} \( \theta \)
\end{itemize}

The unitary operator for the X-axis rotation is given by:
\begin{equation}
U(\theta) = e^{-i \frac{\theta}{2} \sigma_x}
\end{equation}
where \( \sigma_x \) is the Pauli-X operator.

\subsection*{Using Actor-critic Method}

We solve this example by actor-critic method discussed earlier. The actor decides what action to take based on the current state, while the critic evaluates how good the action taken by the actor is by estimating the value function.
Let's see how to integrate it all in RL: 
\begin{itemize}
    \item \textbf{State} \( s \): The quantum state of the qubit, represented as a 2D vector in the Bloch sphere(See fig \ref{blooch}). For example, \( |0\rangle \) is \( [1, 0] \) and \( |1\rangle \) is \( [0, 1] \).
    \item \textbf{Action} \( a \): The rotation angle \( \theta \) applied to the qubit.
    \item \textbf{Reward} \( R(s, a) \): The fidelity between the final state and the target state, defined as:
    \begin{equation}
    R(s, a) = |\langle \psi_{\text{target}} | \psi_{\text{final}} \rangle|^2
    \end{equation}
    \item \textbf{Policy}: The agent's strategy, represented by the actor network, which outputs the optimal rotation angle \( \theta \).
    \item \textbf{Value function}: The critic network estimates the expected future reward for each state.
\end{itemize}

In our quantum control problem, the actor network learns to output the optimal rotation angle \( \theta \), while the critic network estimates the expected cumulative reward (derived from fidelity) for each state. 
The actor and critic are trained using the following rules:
:

\begin{itemize}
    \item The actor's objective is to maximize the expected reward by adjusting \( \theta \). 
    Its loss function is defined as
    \begin{equation}
        L_{\text{actor}} = - A(s,a) \cdot \log \pi_\theta(a|s),
    \end{equation}
    where \( \pi_\theta(a|s) \) is the probability of taking action \(a\) in state \(s\) under the actor policy, and 
    \( A(s,a) \)is the advantage, i.e., the difference between the actual reward and the critic’s value estimate.

    \item The critic is trained to minimize the temporal-difference error:
    \begin{equation}
        L_{\text{critic}} = \big(R_{\text{actual}} - V_{\text{estimated}}\big)^2,
    \end{equation}
    where \( R_{\text{actual}} \) is the reward observed from the environment, and 
    \( V_{\text{estimated}} \) is the value function predicted by the critic.
\end{itemize}


After training the actor and critic networks, we evaluated the learned control strategy by computing the fidelity between the final quantum state and the target state. The fidelity is given by the overlap between the final state \( | \psi_{\text{final}} \rangle \) and the target state \( | \psi_{\text{target}} \rangle \):
\begin{equation}
F = |\langle \psi_{\text{target}} | \psi_{\text{final}} \rangle|^2
\end{equation}

The trained agent successfully learned the optimal rotation angle to drive the qubit from the initial state \( |0\rangle \) to the target state \( |1\rangle \) as shown in the figure \ref{blooch}. The final fidelity achieved was \( F = 0.99999998596 \), indicating a high degree of accuracy in reaching the target state. 

This problem serves primarily as a pedagogical example, and while it effectively illustrates key concepts, employing an actor-critic algorithm in this context may be considered disproportionate given the simplicity of the task.
\subsection{Future Direction: Tracking control in Quantum systems}



Until now, we have discussed RL primarily in the context of quantum optimal control. A growing body of work has demonstrated that RL is highly effective for optimal control tasks such as quantum state preparation \cite{Porotti2022DeepRLQuantum}, gate synthesis \cite{Niu2019}, and dynamical stabilization \cite{Fosel2022RLStabilization}, achieving near-unit fidelities even in complex and noisy settings. 


In most of these studies, the control objective is formulated in terms of a final time target, and the learning agent is rewarded based on a terminal fidelity or cost. As a result, the intermediate system dynamics are optimized only indirectly, insofar as they help reach the desired final state.

There are, however, many physically relevant situations where this formulation is not sufficient. In such cases, the control task requires not only reaching the correct final state, but also regulating the system’s behavior throughout its evolution. This naturally leads to the concept of tracking control, where the objective is to guide the system along a prescribed time dependent trajectory rather than optimizing a terminal outcome.

Early developments in quantum tracking control, particularly in the early 2000s, were based on analytical constructions \cite{Shuang2004,Dong2010}. These approaches were inspired by inverse dynamics and exact tracking methods developed earlier in classical nonlinear control theory \cite{Isidori1995,Marino1995}. The central idea was to derive the control field explicitly from the equations of motion such that a selected observable follows a predefined trajectory. Tracking quantum control has recently attracted a reviewed interest by uncovering novel nonlinear optical effects~\cite{campos_how_2017, mccaul_driven_2020, mccaul_optical_2021, magann_quantum_2023}.

Concretely, consider a controlled quantum system governed by the Hamiltonian
\begin{equation}
H(t) = H_0 + u(t) H_c ,
\end{equation}
where $u(t)$ denotes the control field. If $O$ is the observable of interest, the tracking objective is to enforce
\begin{equation}
y(t) = \langle O(t) \rangle \approx y_d(t),
\end{equation}
where $y_d(t)$ is a prescribed target trajectory. The control task is therefore defined over the full time interval of evolution, rather than at a single final time.

The time derivative of the observable expectation value can be written as
\begin{equation}
\frac{d}{dt}\langle O \rangle
= \frac{i}{\hbar}\langle [H_0, O] \rangle
+ \frac{i}{\hbar} u(t)\langle [H_c, O] \rangle .
\end{equation}
This expression highlights a key structural feature. The observable dynamics depend linearly on the control field. If a desired trajectory $y_d(t)$ is specified, one can impose
\begin{equation}
\frac{d}{dt}\langle O \rangle = \dot{y}_d(t)
\end{equation}
and formally solve for the control field $u(t)$ by inverting the above relation. In this way, analytical tracking enforces the desired trajectory by construction.

While such analytical tracking schemes are elegant and physically transparent, they also reveal important practical limitations. The inversion can become singular, it relies on accurate knowledge of the system Hamiltonian, and it becomes increasingly fragile in the presence of decoherence, noise, or high-dimensional dynamics. Consequently, analytical tracking is most effective for few-level, well characterized systems.

Tracking control can be understood as an inverse-control problem, namely determining the control action required to produce a desired dynamical response. This perspective provides the natural entry point for RL.

In RL-based tracking control, the inverse mapping no longer needs to be written down explicitly. Instead, the control field is generated by a policy,
\begin{equation}
u(t) = \pi\!\left(s(t)\right),
\end{equation}
where the policy $\pi$ maps the agent’s state $s(t)$ to a control action. For tracking problems, the state typically encodes the tracking error,
\begin{equation}
e(t) = y(t) - y_d(t),
\end{equation}
and possibly its recent history or local temporal variation. Rather than enforcing $e(t)=0$ instantaneously, the agent can be trained to reduce the error over time.

This can be achieved by defining a reward signal that penalizes deviations from the desired trajectory, for example,
\begin{equation}
r(t) = -e^2(t).
\label{err}
\end{equation}
The learning objective then becomes the minimization of the accumulated tracking error over the full evolution,
\begin{equation}
\int_0^T e^2(t)\,dt ,
\end{equation}
rather than the optimization of a terminal fidelity. This shift from instantaneous inversion to trajectory level optimization is a defining feature of RL based tracking control. 



Within this setting, several classes of RL algorithms are proposed below, each addressing different aspects of the tracking problem.

A natural starting point is policy-gradient methods such as \cite{sutton1999}, Trust Region Policy Optimization (TRPO) \cite{Schulman2015}, and Proximal Policy Optimization (PPO) \cite{Schulman2017}. In these approaches, the control field is generated directly by a parameterized policy,
\begin{equation}
u(t) = \pi_\theta(s(t)),
\end{equation}
and the policy parameters are updated by maximizing the expected cumulative reward. For tracking problems, the reward is typically defined in terms of the instantaneous tracking error [Eq.~\eqref{err}].
Because policy gradient methods optimize trajectory level objectives without requiring an explicit system model, they are well suited for enforcing time-dependent behavior and smooth control fields. PPO is particularly attractive due to its numerical stability and ease of implementation. Another naturally aligned class of methods is actor-critic reinforcement learning. Algorithms such as DDPG \cite{Lillicrap2015DDPG}, TD3 \cite{Fujimoto2018TD3}, and SAC \cite{Haarnoja2018} may be promising candidates for future tracking control applications. DDPG may be suitable for deterministic tracking control, TD3 could offer improved stability for long-time tracking, and SAC may enable robust tracking by incorporating entropy, which can be beneficial in the presence of noise and uncertainty.  

RL therefore, offers a natural framework for future developments in tracking control of quantum systems, enabling model-free optimization of time-dependent objectives with stable and experimentally feasible control policies.

\section{Quantum Advantage in Learning: Quantum Reinforcement Learning }
\label{QC-QML}

\subsection{Quantum Computing: Introduction}

A quantum circuit is the basic blueprint of how computation happens in a quantum computer. A classical circuit uses wires and logic gates to manipulate bits (0s and 1s) whereas 
a quantum circuit uses quantum gates to manipulate qubits. Let us begin with a simple qubit, which can exist in a combination of two basic states, 
$|0\rangle$ and $|1\rangle$:
\begin{equation}
|\psi\rangle = \alpha |0\rangle + \beta |1\rangle,
\end{equation} 
Quantum gates implement unitary transformations that evolve the quantum state:
\begin{equation}
|\psi'\rangle = U |\psi\rangle, \quad U^\dagger U = UU^\dagger = I.
\end{equation}

Single-qubit gates perform rotations on the Bloch sphere, parameterized as:
\begin{align}
R_j(\theta) &= \exp\left(-i\frac{\theta}{2}\sigma_j\right), \quad j \in \{x, y, z\},
\label{rotation_eq}
\end{align}
where $\sigma_j$ denotes the Pauli matrices. Multi-qubit gates, such as the controlled-NOT (CNOT), create entanglement between qubits, a uniquely quantum resource absent in classical computation.

The measurement process, repeated over many trials (also called shots), yields probability distributions or expectation values:
\begin{equation}
\langle O \rangle = \langle \psi | U^\dagger O U | \psi \rangle,
\label{expt value}
\end{equation}
where $O$ represents an observable operator (e.g., Pauli-$Z$).

In practice, a single measurement of a quantum circuit yields a random outcome, such as \texttt{0110}. 
The power of quantum measurement lies not in one result but in the pattern revealed through many repetitions. 
Each run, or shot, is like a single exposure in photography, unclear alone, but sharp in aggregate. 
By repeating the circuit, we estimate probabilities or expectation values as per Eq.~\eqref{expt value}.
Thus, quantum measurement uncovers information statistically.

\subsection{Quantum Reinforcement Learning }
RL began as a behavioral concept rooted in psychology and control theory, which describes how organisms learn through trial and error by responding to rewards and punishments \cite{Doya2007}. This intuitive idea gradually evolved into a rigorous computational framework in which an agent interacts with an environment to maximize cumulative rewards \cite{Sutton1988,Doya2007}. The modern era of machine learning has reshaped RL into a model-driven approach. Agents are now represented by neural networks and other parameterized systems that learn to make better decisions through experience \cite{Doya2007}. What once described animal learning has become a core framework for developing intelligent, adaptive machines.

This computational framework now extends into the quantum realm. Quantum Reinforcement Learning (QRL) \cite{QRL,Su2025,wang2023quantum} is one of the three principal branches of Quantum Machine Learning (QML), alongside Quantum Supervised and Quantum Unsupervised Learning. Like classical ML, QRL is built around models characterized by trainable parameters—adjustable quantities that enable the system to capture patterns and improve performance through training. A model endowed with such adaptive parameters is known as a learnable model.

In QRL, learnable system is realized through parameterized quantum circuits, where the rotation angles $\theta$ in gates such as $R_x(\theta)$, $R_y(\theta)$, and $R_z(\theta)$ act as trainable parameters. Such circuits, known as Variational Quantum Circuits (VQCs), form the foundation of learnable quantum models. By iteratively tuning $\theta$ via classical optimization, the circuit adapts its behavior to perform computational or decision-making tasks. This concept lies at the heart of Quantum Machine Learning \cite{Ismail2024,Schuld2021}.

To understand quantum RL in more detail, let us revisit policy gradient in Sec.~\ref{Sec11}. 
In the quantum setting, the policy $\pi(a\mid s;\theta)$ is realized by a variational quantum circuit (VQC), and the trainable vector $\theta$ consists of the gates' rotation angles. When the policy is a VQC, the terms $\nabla_\theta \log \pi(a_t\mid s_t;\theta)$ in Eq.~\eqref{reinf}  are obtained on the quantum model, while the optimizer runs classically. Concretely, 
a state is encoded into qubits, the circuit is executed, measurements yield action probabilities, and an action is sampled. After observing returns, those angles are updated using the standard REINFORCE policy-gradient estimator, where gradients are computed via the parameter-shift rule on the quantum circuit and optimization runs classically. In short, the loop is exactly the classical policy-gradient algorithm; the only change is that the function class for $\pi$ is a quantum circuit whose learnable rotation angles form $\theta$.

\section{Conclusion}\label{SecConclusion}

In this tutorial, we have explained several key concepts of Reinforcement Learning (RL) through simple and intuitive examples, making it easy for readers to grasp the foundational topics. Most of the fundamental concepts in RL are covered with straightforward, easy-to-follow tutorials. To complement the theoretical explanations, we have included code implementations that readers can explore. Supporting materials, such as booklets, provide additional guidance, while the complete implementations of the examples discussed in this tutorial are available on GitHub for reference and hands-on practice.

\acknowledgments

A.S. and D.I.B. are supported by Army Research Office (ARO) (grant W911NF-23-1-0288; program manager Dr.~James Joseph). The views and conclusions contained in this document are those of the authors and should not be interpreted as representing the official policies, either expressed or implied, of ARO, or the U.S. Government. The U.S. Government is authorized to reproduce and distribute reprints for Government purposes notwithstanding any copyright notation herein.

\vskip 0.2in
\bibliographystyle{unsrt}
\bibliography{ref_new}

\appendix

\section{Bellman Equation Derivation}\label{bellman-details}

The Bellman expectation equation, introduced in Sec. \ref{objective2} [Eq.~\ref{bellman}], plays a central role in the analysis presented in this work. For completeness, its derivation is outlined in this appendix, starting from the fundamental definitions of the return and the state-value function introduced earlier.

The return at time step $t$, defined in Eq.\ref{Discount_factor} is given by
\begin{equation}
G_t = \sum_{k=0}^{\infty} \gamma^k r_{t+k+1},
\end{equation}
where $\gamma \in [0,1)$ is the discount factor. This definition satisfies the recursive relation
\begin{equation}
G_t = r_{t+1} + \gamma G_{t+1}.
\end{equation}

The state-value function under a fixed policy $\pi$, defined in Eq.\ref{value_function_stochastic}, is
\begin{equation*}
v_\pi(s) = \mathbb{E}_{\pi}\!\left[ G_t \mid s_t = s \right].
\end{equation*}
Substituting the recursive form of the return into the above definition yields
\begin{equation}
v_\pi(s)
=
\mathbb{E}_{\pi}\!\left[ r_{t+1} + \gamma G_{t+1} \mid s_t = s \right].
\end{equation}

Using the linearity of expectation, the expression can be separated into immediate and future contributions,
\begin{equation*}
v_\pi(s)
=
\mathbb{E}_{\pi}\!\left[ r_{t+1} \mid s_t = s \right]
+
\gamma
\mathbb{E}_{\pi}\!\left[ G_{t+1} \mid s_t = s \right].
\end{equation*}

The expectation over future rewards can be expanded by conditioning on the action $a_t$ selected according to the policy $\pi(a|s)$,
\begin{equation*}
v_\pi(s)
=
\sum_{a} \pi(a|s)
\mathbb{E}\!\left[ r_{t+1} + \gamma G_{t+1}
\mid s_t = s, a_t = a \right].
\end{equation*}

Further conditioning on the next state $s_{t+1} = s'$ reached under the environment dynamics leads to

\begin{equation}
\begin{aligned}
v_\pi(s)
&=
\sum_{a} \pi(a\mid s)
\sum_{s'} P(s'\mid s,a)\,
\mathbb{E}\!\Bigl [
r_{t+1} + \gamma G_{t+1}
\\
&\qquad\qquad\qquad
\mid s_t = s,\ a_t = a,\ s_{t+1} = s'
\Bigr].
\end{aligned}
\end{equation}

Note that we add $\sum_{a}$ because the policy randomly chooses actions, and further add $\sum_{s'}$ because the environment
randomly transitions to the next state, therefore, conditioning averages over both sources of randomness.
In other words, we average over both the policy and the environment because $v_\pi(s)$ is an expectation and  the return is random due to (i) stochastic action selection under $\pi(a\mid s)$ and (ii) stochastic state transitions under $P(s'\mid s,a)$. Consequently, computing $v_\pi(s)$ requires taking the expectation over both sources of randomness.

Assuming Markovian dynamics, the immediate reward depends only on the current state, action, and next state,
\begin{equation}
\mathbb{E}\!\left[ r_{t+1} \mid s,a,s' \right] = R(s,a,s'),
\end{equation}
while the expected future return from time $t+1$ onward depends only on the next state,
\begin{equation}
\mathbb{E}\!\left[ G_{t+1} \mid s_{t+1} = s' \right] = V^{\pi}(s').
\end{equation}

Substituting these relations into the previous expression yields
\begin{equation}
v_\pi(s)
=
\sum_{a} \pi(a|s)
\sum_{s'} P(s'|s,a)
\left[
R(s,a,s') + \gamma V^{\pi}(s')
\right].
\end{equation}

This expression corresponds to the Bellman expectation equation given in Eq.\ref{bellman}.

\section{Bellman Optimality Condition}\label{bellappen}

In the Sec.~\ref{objective2}, we introduced the Bellman equation. We learned that Bellman equation Eq (\ref{bellman}) helps us in evaluating the value of a state $v_{\pi}(s)$ under the policy $\pi$. The Bellman optimality equation is an extension of the Bellman equation. While the standard Bellman equation expresses the value of a state \( v_\pi(s) \) under a given policy \(\pi\) as

\[
v_\pi(s) = \sum_a \pi(a \mid s) \sum_{s',r} p(s', r \mid s, a) \left[ r + \gamma\, v_\pi(s') \right],
\]

the Bellman optimality equation describes the value of a state under the optimal policy—that is, the policy that maximizes the expected return. \textbf{Note:} We reach the optimal policy through a process known as \emph{policy improvement}. Starting with an initial policy \(\pi\), we iteratively update the policy to obtain \(\pi_1\), \(\pi_2\), \(\dots\), until we converge to the optimal policy, which we denote by \( * \) (can also denote it as \(\pi_*\)). As we iteratively improve the policy, the state values \( v_{\pi}(s) \) for each state \( s \) effectively increases. This process of policy improvement ensures that we converge towards the optimal policy, where the state values represent the maximum expected return.

In the optimal case, the equation is given by

\begin{equation}
v_*(s) = \max_a \sum_{s',r} p(s', r \mid s, a) \left[ r + \gamma\, v_*(s') \right],
\label{bellman_optimality}
\end{equation}

where \(v_*(s)\) denotes the optimal value of state \(s\).

Let us understand the above equation. For simplicity, let us consider the following situation: suppose that the agent is at \(s_{4}\), as shown in Figure \ref{fig:improvedpolicy}. Now, at this state, the action set \( \mathcal{A} \) is given by\[
\mathcal{A} = \{\texttt{left}, \texttt{right}\}, \quad a \in \mathcal{A},
\]
Let us now calculate the two value functions $v^{left}_\pi(s_4)$ and $v^{right}_\pi(s_4)$ corresponding to the two possible actions left and right respectively.

Using the Bellman expectation equation (note that since we define the action explicitly, the summation over the policy probabilities \(\pi(a|s_4)\) disappears)):
\[
v^{left}_\pi(s_4) = \sum_{s',r} p(s', r \mid s_4, left) \left[ r + \gamma v_\pi(s') \right]
\]

Given the transition probabilities:
\[
p(s' \neq s_3, r \mid s_4, left) = 0
\]

This simplifies the value function to:
\[
\begin{aligned}
v^{left}_\pi(s_4) &= p(s' = s_3, r = -1 \mid s_4, left) \left[ -1 + \gamma v_\pi(s_3) \right] \\
&= p(s' = s_3, r = -1 \mid s_4, left) \times \left( -1 + \gamma v_\pi(s_3) \right)
\end{aligned}
\]

Similarly,
 \vspace{-0.6em}
\[
v^{right}_\pi(s_4) = p(s' = s_5, r = 3 \mid s_4,right) \left[ 3 + v_\pi(s_5) \right]
\]
Now for the shake of simplicity let us assume the following
\[
p(s' = s_5, r = 3 \mid s_4,  right) = 1
\]
\vspace{-2.5em}
\[
p(s' = s_3, r = -1 \mid s = s_4, a = left) = 1
\]
which leads to

\begin{equation}
\begin{aligned}
v_{\pi}^{left}(s_4) &= \left[ -1 + v_\pi(s_3) \right] \\
v_{\pi}^{right}(s_4) &= \left[ 3 + v_\pi(s_5) \right] \\
\text{where } v_\pi(s_3) &\leq 3 \text{ (assume } v_\pi(s_3)\leq \text{ $r_{max}$)} \\
\text{and } v_\pi(s_5) &= 0 \text{ ( as } s_5 \text{ is optimal state)} \\
\text{Thus, } v_{\pi}^{left}(s_4) &\leq \left[ -1 + 3 \right] = 2 \\
v_{\pi}^{right}(s_4) &= \left[ 3 + 0 \right] = 3
\label{final_key}
\end{aligned}
\end{equation}


According to the Bellman optimality condition, the optimal state value at \( s_4 \) is obtained by taking the maximum over all possible actions. That is, we have:
\[
v_{*}(s_4) = \max \{ v_\pi^{left}(s_4), \, v_\pi^{right}(s_4) \}.
\]

From our earlier calculations, we know that:
\[
v_\pi^{left}(s_4) \leq 2 \quad \text{and} \quad v_\pi^{right}(s_4) = 3.
\]

Therefore, the optimal state value is:
\[
v_{*}(s_4) = \max \{ 2, 3 \} = 3.
\]

This result indicates that the \texttt{right} action is optimal at \( s_4 \) (i.e., \( a^* = \texttt{right} \)) because it yields the highest expected return. In summary, by applying the Bellman optimality equation, we determine that choosing the \texttt{right} action maximizes the value function at state \( s_4 \).

\section{Iterative Method}\label{AppedixIterMeth}

In this section, we will show how the Bellman equation, which was solved analytically in section \ref {objective2}, can  be solved iteratively.  

 Before we show this, let us briefly recapitulate what an iterative solution method is. Consider the following equation.
 \[
x = f(x)
\]

An iterative method aims to find the solution to this equation by starting from an initial guess and refining it step-by-step. The process is defined as:  
\[
x_{k+1} = f(x_k)
\]

where:  
\begin{itemize}
    \item \(x_k\) is the approximation of the solution at iteration \(k\).  
    \item \(f(x)\) is the function that updates the current estimate.  
\end{itemize}

The iterations continue until the solution converges, i.e., until the difference between consecutive estimates is below a predefined threshold (the stopping criterion):
\[
|x_{k+1} - x_k| < \epsilon
\]
Let \( f(x) = \cos(x) \). Starting with \( x_0 = 0 \) and using the iteration formula \( x_{\text{new}} = \cos(x_{\text{old}}) \), the first three iterations are:
\[
\begin{aligned}
x_0 &= 0, \\
x_1 &= \cos(x_0) = \cos(0) = 1, \\
x_2 &= \cos(x_1) = \cos(1) \approx 0.5403, \\
x_3 &= \cos(x_2) = \cos(0.5403) \approx 0.8576.
\end{aligned}
\]

Continuing the iterations until the stopping criterion is satisfied, we stop after the 10th iteration. At this point, the obtained solution is
\[
x \approx 0.7314.
\] 

In the next section, we will present the iterative method in the case of the Bellman equation.

\section{Iterative policy evaluation}\label{AppedixIterPolicy}

In Sec.~\ref{objective2}, we obtained the value functions for different states by solving a system of linear equations Eq.~\eqref{eq20} - \ref{23}. However, the same can be done using iterative method discussed in previous section. This iterative method of policy evaluation follows the equation:

\begin{equation}
\begin{aligned}
v^{k+1}_{\pi}(s) &= \sum_{a} \pi(s | a) \sum_{s', r} p(s', r \mid s, a) \left[ r + \gamma\, v^k_{\pi}(s') \right]
\label{queen}
\end{aligned}
\end{equation}

Now in principle
\begin{equation}
\lim_{k \to \infty} v^{k}_{\pi}(s) = v_{\pi}(s) \quad \text{for any } s \in S
\label{master_eqn}
\end{equation}
where $v_{\pi}(s)$ is the values function obtained from the solution of system of linear equation obtained under the specific policy as discussed in Sec.~\ref{objective2}.

For instance, let's consider the task of calculating the value function \( v_\pi(s_2) \) for the state \( s_2 \) (see Figure \ref{fig:improvedpolicy}) under a given policy, denoted as \( \pi \), which assigns probabilities to actions taken at state \( s_2 \). This policy is defined as:

\[
\pi: \{ \texttt{left}: p_1, \texttt{right}:  p_2 \}
\]

where \( p_1 \) and \( p_2 \) represent the probabilities of the agent taking the \texttt{left} and \texttt{right} actions, respectively. 
From \eqref{master_eqn} we have
\[
\lim_{k \to \infty} v^k_{\pi}(s_2) = v_{\pi}(s_2)
\]
Now the task is to calculate $v_{k}(s_2)$ using the iterative method.
From \eqref{queen} the iterative value update equation of  \( v_{\pi}^{l+1}(s_2) \) is:
\[
\forall l \in \mathbb{Z}^+ \text{ such that } l + 1 < k \in \mathbb{Z}^+
\]
Therefore, 
\begin{equation}
\begin{aligned}
 v_{\pi}^{l+1}(s_2) &= \sum_{a} \pi(s_2 | a) \sum_{s', r} p(s', r \mid s_2, a) \left[ r + \gamma\, v_{\pi}^l(s') \right] \\
 &= p_1 p(s_3, r \mid s_2, a) \left[ r_3 + \gamma v_{\pi}^l(s_3) \right] \\
 &\quad + p_2 p(s_1, r \mid s_2, a) \left[ r_1 + \gamma v_{\pi}^l(s_1) \right] \\
 &= p_1 \left[ r_3 + \gamma v_{\pi}^l(s_3) \right] + p_2 \left[ r_1 + \gamma v_{\pi}^l(s_1) \right]
 \label{mine_eq}
\end{aligned}
\end{equation}
where, in above derivation, we have used
$\quad p(s_3, r \mid s_2, a) = 1 =  p(s_1, r \mid s_2, a). $ \\

In order to calculate \( v_{\pi}^{l+1}(s_2) \), it is necessary to have the values of \( v_{\pi}^l(s_3) \) and \( v_{\pi}^l(s_1) \) from the previous iteration. These values can be obtained through an iterative method that updates the value functions \( v_{\pi}^l(s) \) for all states \( s \in S \) at the \( l \)-th iteration. The updated values are then utilized to compute the value function for the \( (l+1) \)-th iteration.

The initialization of the value functions can be chosen as follows:

\[
v_{\pi}^{l=0}(s) = 0 \quad \text{for all states } s \in S.
\].

In practice, zeros are often used because they are simple and serve as a neutral starting point, but if one has prior knowledge about the value function,  it is better to choose a different initialization to potentially speed up convergence.

\section{Value iteration}
Value iteration addresses the drawback of policy iteration by combining policy evaluation and improvement into a single step, performing just one update per state rather than fully evaluating the policy. This streamlined approach uses the Bellman optimality equation to ensure convergence while reducing computational cost.
The equation that governs the value iteration is the following,
the equation
\begin{equation}
\begin{aligned}
 v^{k+1}(s) = \max_{a} \sum_{s',r} p(s',r|s,a) \left[ r + \gamma v^k(s') \right]
\end{aligned}
\end{equation}
Now here we will also have the following relation
\[
\lim_{k \to \infty} v^k(s) = v_{*}(s)
\]
The value iteration equation above can also be understood in a different way: as turning the Bellman optimality equation into an update rule. Instead of performing a full policy evaluation for a fixed policy, it directly incorporates policy improvement by selecting, at each state, the action that maximizes the expected return. This streamlined approach combines evaluation and improvement into a single update step.

\section{First visit and Every visit Monte Carlo}\label{AppenixFirsEveryZVisit}

\textbf{First visit:}
When an agent encounters a state \( s \) during an episode, it will record the return (sum of rewards from that point onwards) the first time it visits that state. The state’s value is then updated based on that return. If the state is visited again later in the same episode, it is ignored for the purpose of updating the value estimate.
\[
v(s) = \frac{1}{N(s)} \sum_{t \in \text{First Visit to } s}G_t
\]
N(s) \text{ is the number of first visits to state } s,\\ 
\quad $G_t$ \text{ is the return starting from time step } t.
\vspace{0.15cm}
\\

\textbf{Every visit: }
Whenever the agent visits a state \( s \), the value of that state is updated based on the return \( G_t \) (the total reward from that point onward in the episode). The value of the state is updated multiple times within an episode if it is visited multiple times.
Let the return for state \( s \) at time \( t \) be \( G_t \), and the number of times state \( s \) has been visited up to that point be \( N(s) \). The value of state \( s \), \( V(s) \), is updated as:

\[
V(s) = \frac{1}{N(s)} \sum_{t \in \text{visits to } s} G_t
\]

\end{document}